\documentclass{article}

\usepackage[preprint]{neurips_2026}

\usepackage[utf8]{inputenc}
\usepackage[T1]{fontenc}
\usepackage{xcolor}
\definecolor{MyBlue}{RGB}{0,172,238}
\definecolor{MyPink}{RGB}{237,0,139}
\usepackage[colorlinks=true,
            linkcolor=red,
            citecolor=MyBlue,
            urlcolor=MyPink
           ]{hyperref}
\usepackage{url}
\usepackage{booktabs}
\usepackage{amsfonts}
\usepackage{nicefrac}
\usepackage{microtype}
\usepackage{xcolor}

\usepackage{amsmath}
\usepackage{graphicx}
\usepackage{multirow}
\usepackage{stackrel}
\usepackage{enumitem}
\usepackage{bm}

\usepackage{colortbl}
\usepackage{ulem}
\usepackage{multirow}
\usepackage{makecell}
\usepackage{subcaption}
\usepackage[dvipsnames]{xcolor}
\usepackage[most]{tcolorbox}
\usepackage{xcolor}
\usepackage{pifont}
\usepackage{placeins}
\newcommand{\insightbox}[1]{
    \begin{tcolorbox}[
        colframe=black!70,
        colback=yellow!5,
        boxrule=1pt,
        arc=4mm,
        width=\textwidth,
        left=4mm,
        right=4mm,
        top=0.2mm,
        bottom=1mm
    ]
        \raisebox{-0.2cm}{\includegraphics[width=0.5cm]{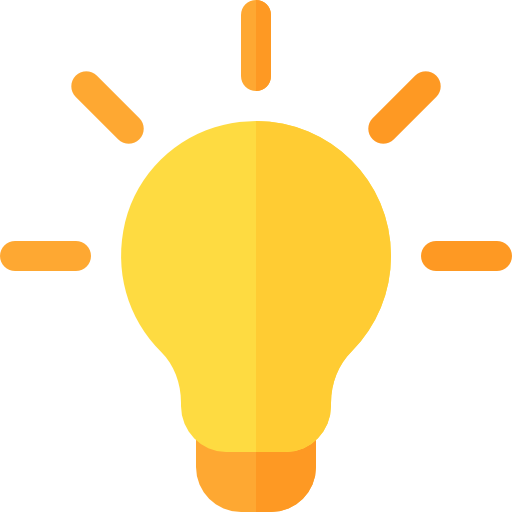}}
        \textbf{\small #1}
    \end{tcolorbox}
}
\usepackage{bm}
\usepackage{graphicx}
\usepackage{booktabs}
\usepackage{pifont}
\usepackage{tabularx}
\usepackage{booktabs}
\usepackage{multirow}
\usepackage{float}
\usepackage{orcidlink}
\usepackage{graphicx}
\usepackage{tcolorbox}
\usepackage{graphicx}
\usepackage{multirow}
\usepackage[table]{xcolor}
\usepackage{makecell}
\usepackage{multirow}
\usepackage{wrapfig}

\usepackage{algpseudocode}
\usepackage{algorithmicx}
\usepackage[dvipsnames]{xcolor}
\usepackage[ruled,vlined]{algorithm2e}

\SetCommentSty{mycommfont}

\title{PathMem: Toward Cognition-Aligned Memory Transformation for Pathology MLLMs}

\author{%
  Jinyue Li\textsuperscript{1,$\ast$} \quad
  Yuci Liang\textsuperscript{2,$\ast$} \quad
  Qiankun Li\textsuperscript{3, 4$\dagger$} \quad
  Xinheng Lyu\textsuperscript{2} \quad
  Jiayu Qian\textsuperscript{3} \quad
  Huabao Chen\textsuperscript{1} \\
  \textbf{Kun Wang}\textsuperscript{3} \quad
  \textbf{Zhigang Zeng}\textsuperscript{5} \quad
  \textbf{Anil Anthony Bharath}\textsuperscript{4} \quad
  \textbf{Yang Liu}\textsuperscript{3} \\
  \\
  \textsuperscript{1}University of Science and Technology of China \quad
  \textsuperscript{2}Shenzhen University \\
  \textsuperscript{3}Nanyang Technological University \quad
  \textsuperscript{4}IGS, Imperial College London\\
  \textsuperscript{5}Huazhong University of Science and Technology
}

\begin{document}

\maketitle

\begin{abstract}
Computational pathology demands both visual pattern recognition and dynamic integration of structured domain knowledge, including taxonomy, grading criteria, and clinical evidence.
In practice, diagnostic reasoning requires linking morphological evidence with formal diagnostic and grading criteria.
Although multimodal large language models (MLLMs) demonstrate strong vision–language reasoning capabilities, they lack explicit mechanisms for structured knowledge integration and more reliable memory control.
As a result, existing models struggle to consistently incorporate pathology-specific diagnostic standards during reasoning.
Inspired by the hierarchical memory process of human pathologists, we propose \texttt{PathMem}, a memory-centric multimodal framework for pathology MLLMs.
\texttt{PathMem} organizes structured pathology knowledge as a long-term memory (LTM) and introduces a \textbf{Memory Transformer} that models the dynamic transition from LTM to working memory (WM) through multimodal memory activation and context-aware knowledge grounding, enabling context-aware memory refinement for downstream reasoning.
\texttt{PathMem} achieves SOTA performance across benchmarks, improving WSI-Bench report generation (\textbf{+12.8\%} WSI-Precision, \textbf{+10.1\%} WSI-Relevance) and open-ended diagnosis by \textbf{+9.7\%} and \textbf{+8.9\%} over prior WSI-based models.
\end{abstract}

\begin{figure*}[!h]
\centering
\includegraphics[width=1\textwidth]{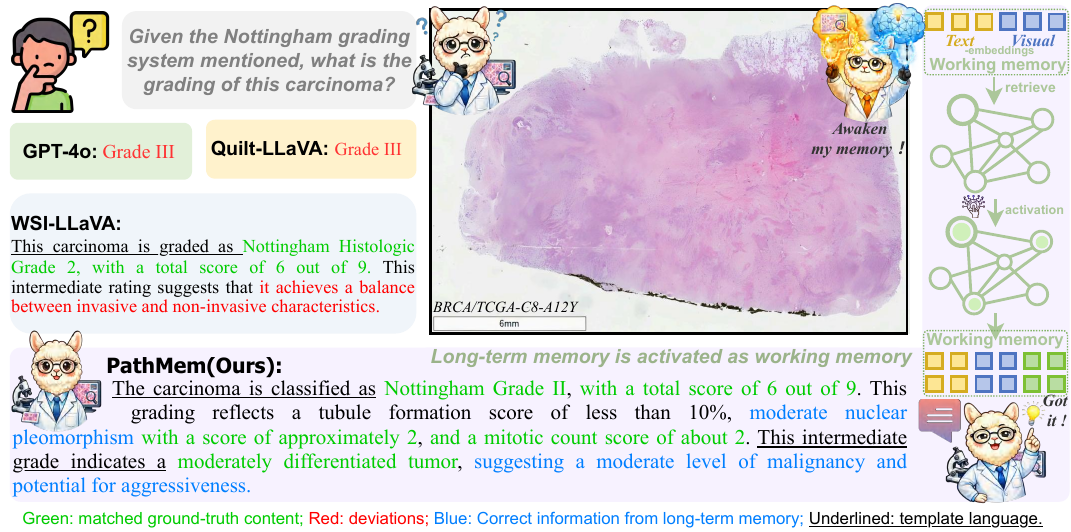}
\caption{
Overview of \texttt{PathMem}. The working memory selectively activates relevant long-term memory traces and transforms them into an updated working memory state for more reliable reasoning.
}
\label{Figure1}
\end{figure*}
\section{Introduction}
\label{sec:Introduction}

Computational pathology is a knowledge-intensive and cognitively complex discipline \cite{omar2024chatgpt,liu2026benchmarking,cifci2023ai,liang2025wsi,sun2025cpath,ahmed2024pathalign,lu2024multimodal,chen2025evidence}. Accurate diagnosis requires not only visual pattern recognition from histopathological images but also the dynamic integration of long-term expert knowledge~\cite{verghese2023computational}, including disease taxonomy~\cite{shafique2024preliminary,redline2023placental,romero2022toward,gabitto2024integrated}, grading criteria~\cite{jiang2024transformer,gosai2025beyond,li2025multi}, and evolving clinical evidence~\cite{fountzilas2025convergence,ullah2025multimodal,bian2026biomedical}.
In practice, diagnostic reasoning requires linking morphological observations with formal diagnostic and grading standards.
Although recent multimodal large language models (MLLMs)~\cite{yin2024survey,caffagni2024revolution,liu2023visual} have demonstrated promising vision-language reasoning capabilities, they largely operate as parametric black boxes and lack explicit mechanisms for structured knowledge integration and more reliable memory control~\cite{liu2025unveiling,cabalo2024differential,sheng2025top}.
Therefore, incorporating structured, controllable, and more reliable knowledge integration mechanisms into multimodal reasoning architectures may become a key challenge for building the next generation of clinically reliable and cognitively robust intelligent systems in computational pathology.

In clinical practice, pathologists rely on a hierarchical memory process \cite{hou2024memory,hu2025loc,liu2025pathmind,chen2025pathagent,sun2024pathmmu}: accumulated domain knowledge forms long-term memory (LTM)~\cite{bei2026mem,yang2022long,minhas2024restoring,blackmore2023ultrasound}, while case-specific evidence selectively activates relevant knowledge into working memory (WM) for reasoning \cite{baddeley2025working,akhtar2024types,karakatsani2023focused,manchanda2023intravenous}, and iteratively refine their hypotheses. Existing retrieval-augmented generation (RAG) methods attempt to introduce external knowledge \cite{zhang2025patho,hamza2025llava,jabal2025open,omar2024chatgpt,yu2025ypathrag,fink2025retrieval}, but their retrieval pipelines are typically static and fail to model the dynamic and adaptive nature of human memory transformation. This cognitive process, characterized by memory retrieval, selective activation, and meta-cognitive regulation, is largely absent in current MLLMs \cite{an2025cognitive,huang2026ama,jiang2026synapse,hou2025synapticrag,yoshida2025if,dong2025towards}.

In this paper, we propose \texttt{PathMem}, a memory-centric multimodal architecture for pathology MLLMs (\textcolor{red}{Figure \ref{Figure1}}). We first construct a \textbf{high-quality pathology knowledge graph} as LTM via deep semantic search over PubMed, augmented with diverse sources, including clinical guidelines, gray literature, among others.
The knowledge graph is encoded into semantic embeddings aligned with the multimodal backbone.
We then design a \textbf{Memory Transformer} that explicitly models the transition from LTM to WM. Given visual and textual embeddings, we perform:
Static activation, which ranks knowledge entries via cosine similarity;
Dynamic activation, which jointly projects multimodal and knowledge embeddings to compute global relevance.
An adaptive selection strategy determines the boundary of activated knowledge, transferring only highly relevant entries into WM for final reasoning. This mechanism enables controlled and more reliable memory enhancement during inference.
In report generation and open-ended diagnosis, it improves WSI-Precision / WSI-Relevance by \textbf{12.8\%} / \textbf{10.1\%} and \textbf{9.5\%} / \textbf{8.9\%}, respectively.

    The contributions of this paper are summarized as follows.
 \begin{itemize}
 	\item[\ding{182}] \textbf{Memory-Driven Pathology MLLM Architecture}: We propose \texttt{PathMem}, an explicit long-term/working memory paradigm into multimodal pathology architecture, enabling cognitively grounded and knowledge-aware reasoning.
	\item[\ding{183}] \textbf{High-Quality LTM Construction}: We build a structured pathology knowledge graph via deep search over PubMed, providing a scalable and updatable repository that simulates accumulated expert-level domain knowledge.
    \item[\ding{184}] \textbf{Dynamic–Static Memory Controller with Adaptive Activation}: We propose a dual-mode memory activation mechanism with self-adaptive selection, explicitly modeling the transformation from LTM to WM for context-aware, more reliable, and enhanced inference.
    \item[\ding{185}] \textbf{PathMem achieves the SOTA performance across multiple Benchmarks}:
    In report generation, it improves WSI-Precision by \textbf{12.8\%} and WSI-Relevance by \textbf{10.1\%}. In open-ended diagnosis, it further yields gains of \textbf{9.5\%} and \textbf{8.9\%}, respectively.
\end{itemize}
\insightbox{
By replacing static retrieval with explicit memory transformation, our framework enables more reliable, memory-aware reasoning aligned with human cognition in pathology MLLMs.
}

\section{Related Work}
\label{sec:Related Work}

\paragraph{Pathology MLLM.}
With the rapid development of computational pathology, the availability of large-scale whole slide images (WSIs) paired with diagnostic reports has catalyzed the rise of pathology-oriented MLLMs~\cite{xu2024whole, ding2025multimodal, ahmed2024pathalign, zhou2025mllm4pue, chen2024metapath,allaberdiev2026chestxgen,chen2025evidence}.
For instance, Prov-GigaPath~\cite{xu2024whole} and TITAN~\cite{ding2025multimodal} emphasize large-scale whole-slide modeling and vision–language alignment, while PRISM~\cite{shaikovski2024prism} and PathAlign~\cite{ahmed2024pathalign} demonstrate the effectiveness of language-aligned slide embeddings for zero-shot and generative applications.
Building upon these slide-level foundations, recent pathology MLLMs further integrate instruction tuning, task unification, and computational efficiency into a cohesive framework. PathAsst~\cite{sun2024pathasst} and CPath-Omni~\cite{sun2025cpath} construct unified multimodal assistants that consolidate patch- and slide-level tasks under large-scale image–text supervision. SlideChat~\cite{chen2025slidechat} and WSI-LLaVA~\cite{liang2025wsi} enhance morphology-aware reasoning and conversational capabilities through dedicated WSI-benchmarks and instruction data. To address the prohibitive cost of gigapixel modeling, LoC-Path~\cite{hu2025loc} introduces redundancy-aware compression for efficient slide representation, while ChatEXAONEPath~\cite{kim2025chatexaonepath} strengthens expert-level clinical reasoning via retrieval-based data construction.
\ding{168} In summary, while \textit{prior pathology MLLMs achieve strong slide-level understanding through large-scale pretraining and alignment, they lack explicit mechanisms to balance long-term knowledge and short-term contextual memory, limiting stable and consistent clinical reasoning.}

\paragraph{Knowledge in pathology.}
Explicit domain knowledge is essential for robust computational pathology \cite{ma2025generalizable,smart1995model,huang2025knowledge,jiang2026pathreasoner,zhao2024aligning}.
Knowledge has been incorporated as an inductive bias during pretraining and adaptation. Unified knowledge distillation aggregates expert and self-knowledge to enhance cross-task generalization \cite{ma2025generalizable}, while knowledge-guided adaptation leverages visual and textual priors to improve cross-domain robustness and demographic fairness \cite{huang2025knowledge}. These works highlight the role of knowledge in stabilizing and generalizing foundation models.
Beyond improving robustness, another line of work aligns WSIs with explicit expert concepts. Knowledge concept-based MIL frameworks connect slide representations with disease-specific linguistic concepts derived from medical literature \cite{zhao2024aligning}, enabling semantically grounded modeling. In multimodal prognosis prediction, prior and textual knowledge guide phenotype extraction and genotype interaction \cite{he2025prior, shi2025kpvg}, reducing modality gaps and enhancing clinically meaningful integration.
Recently, knowledge has been injected into pathology vision-language models to promote structured reasoning. By aligning WSIs with medical knowledge graphs and optimizing reasoning trajectories, knowledge-guided policy learning enables evidence-linked, chain-of-thought diagnosis \cite{jiang2026pathreasoner}, improving interpretability beyond outcome-level prediction.
Despite progress, most knowledge-enhanced methods rely on implicit supervision or proprietary RAG pipelines, limiting reproducibility and long-term knowledge utilization.
\ding{168} Different from \textbf{prior knowledge-injection} strategies that treat knowledge as auxiliary supervision or external retrieval,
\textit{we approach enhancement from a \textbf{memory-aware perspective}
by constructing a high-quality pathology knowledge base and integrating it as structured LTM within pathology MLLMs, enabling persistent, explicit, and more reliable reasoning.}

\section{Method}
\label{Method}

\subsection{Unified Model Interface}
\textbf{Problem Formulation.}
We formulate LTM construction as a literature-grounded knowledge aggregation problem.
Given a domain-specific query $q_0$, we construct a weighted directed multigraph
$\mathcal{G}$ encoding disease–feature–evidence relations extracted from biomedical abstracts.
Formally, the resulting LTM is defined as
\begin{equation}
\mathcal{G}
=
\left(
\bm{\mathcal{V}},
\bm{\mathcal{R}},
\bm{\mathcal{E}},
\mathbf{W},
\Phi,
\Psi
\right),
\end{equation}
where $\mathcal{V}$ denotes the global entity set, $\bm{\mathcal{R}}$ denotes the predefined relation schema,
$\bm{\mathcal{E}} \subseteq \bm{\mathcal{V}} \times \bm{\mathcal{R}} \times \bm{\mathcal{V}}$ denotes the set of directed relational triples,
$\mathbf{W} : \bm{\mathcal{E}} \rightarrow [0,1]$ assigns calibrated confidence weights to edges,
$\Phi$ denotes the entity normalization mapping, and $\Psi$ denotes the feature-based inverted index for retrieval.
The objective is to maximize structured knowledge coverage while controlling redundancy and uncertainty propagation.
\begin{figure*}[!h]
\centering
\includegraphics[width=\textwidth]{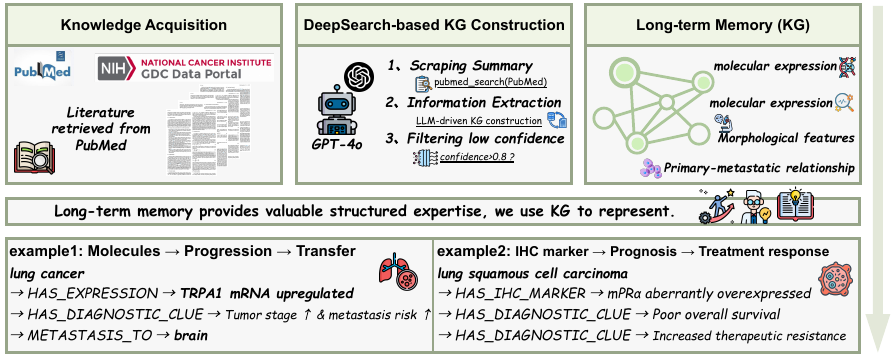}
\caption{LTM construction pipeline for pathology knowledge graphs via iterative literature retrieval and LLM-based information extraction.}
\label{KGV3}
\end{figure*}
\subsection{LTM Construction}
As illustrated in \textcolor{red}{Figure~\ref{KGV3}}, we construct the LTM through an iterative pipeline that integrates literature retrieval with LLM-based information extraction to progressively build and refine the pathology knowledge graph.
\textbf{\ding{168} Hash-Based Deduplication and Memory Monotonicity.}
To guarantee monotonic growth of memory without redundancy,
we define a deterministic hashing operator $\bm{\bm{\mathcal{H}}(\cdot)}$ and compute
$h_i
=
\bm{\mathcal{H}}
\left(
\mathrm{Normalize}(a_i)
\right),
$
where $\mathrm{Normalize}(\cdot)$ removes formatting noise, punctuation variation,
and case differences before hashing.
Let $\bm{\bm{\mathcal{M}}}_h^{(t)}$ denote the set of previously observed hash values at iteration $t$.
The deduplicated abstract set is then defined as
\begin{equation}
\bm{\mathcal{A}}^{*}
=
\left\{
a_i
\;\middle|\;
h_i \notin \bm{\mathcal{M}}_h^{(t)}
\right\},
\quad
\bm{\mathcal{M}}_h^{(t+1)}
=
\bm{\mathcal{M}}_h^{(t)}
\cup
\left\{
h_i
\right\},
\end{equation}
here $h_i$ is the hash signature of abstract $a_i$,
$\bm{\mathcal{M}}_h^{(t)}$ is the memory of previously seen abstracts,
and $\bm{\mathcal{A}}^{*}$ is the effective corpus after deduplication.
This guarantees that identical textual evidence does not repeatedly influence the long-term graph.
\textbf{ \ding{165} LLM-Based Structured Knowledge Extraction.}
Each deduplicated abstract $a_i \in \bm{\mathcal{A}}^{*}$ is processed by a large language model extractor
$f_{\theta}$ parameterized by $\theta$:
$
\mathcal{T}_i
=
f_{\theta}(a_i)
=
\left\{
\left(
s_{ij},
r_{ij},
o_{ij},
c_{ij},
\mathbf{z}_{ij}
\right)
\right\}_{j=1}^{K_i},
$
here $K_i$ is the number of candidate triples extracted from abstract $a_i$,
$s_{ij} \in \bm{\mathcal{V}}$ is the subject entity,
$r_{ij} \in \bm{\mathcal{R}}$ is the relation type,
$o_{ij} \in \bm{\mathcal{V}}$ is the object entity,
$c_{ij} \in [0,1]$ is the model-estimated confidence score,
and $\mathbf{z}_{ij} \in \mathbb{R}^{d}$ is the latent semantic embedding of the triple,
capturing contextual representation for downstream similarity consolidation.
\textbf{\ding{68} Confidence-Constrained Triple Filtering.}
To ensure high precision in LTM,
we introduce a minimum confidence threshold $\tau$.
The retained triple set from abstract $a_i$ is defined as
\begin{equation}
\bm{\mathcal{T}}_i^{\tau}
=
\left\{
\left(
s_{ij},
r_{ij},
o_{ij},
c_{ij},
\mathbf{z}_{ij}
\right)
\in
\bm{\mathcal{T}}_i
\;\middle|\;
c_{ij} \ge \tau
\right\},
\end{equation}
here $\tau \in (0,1)$ controls the trade-off between recall and precision,
and is empirically determined on a validation set to balance precision and coverage.
Only triples whose estimated confidence exceeds $\tau$ are injected into the knowledge graph.
\textbf{\ding{171} Probabilistic Multi-Evidence Fusion.}
Since identical semantic triples may appear across multiple abstracts,
we perform probabilistic evidence aggregation.
For a canonical triple $(s,r,o)$ appearing in $m$ independent sources,
with confidence scores $\{c^{(1)}, c^{(2)}, \dots, c^{(m)}\}$,
the aggregated edge weight is defined as
\begin{equation}
\bm{\mathcal{W}}(s,r,o)
=
1
-
\prod_{k=1}^{m}
\left(
1
-
\alpha
\cdot
c^{(k)}
\cdot
\exp
\left(
-
\bm{\mathcal{F}}
\,
\left\|
\mathbf{z}^{(k)}
-
\bar{\mathbf{z}}
\right\|_2^2
\right)
\right),
\end{equation}
where $\alpha \in (0,1]$ is a global scaling coefficient,
$c^{(k)}$ is the confidence score from the $k$-th abstract,
$\mathbf{z}^{(k)}$ is the embedding representation of the $k$-th evidence instance,
$\bar{\mathbf{z}}$ is the centroid embedding of all supporting instances,
$\|\cdot\|_2$ denotes Euclidean norm,
and $\bm{\mathcal{F}} > 0$ controls embedding-consistency penalization.
This formulation follows a noisy-or probabilistic fusion mechanism,
where multi-source corroboration increases edge certainty,
while embedding inconsistency reduces effective contribution.
\textbf{\ding{169} Feature-Oriented Memory Indexing.}
To support efficient retrieval during reasoning,
we construct a feature-inverted index
$
\bm{\Psi}(f)
=
\left\{
(s,r,o)
\in
\bm{\mathcal{E}}
\;\middle|\;
f = s
\;\lor\;
f = o
\right\},
$
where $f \in \bm{\mathcal{V}}$ denotes a histopathological feature entity,
and $\Psi(f)$ returns all triples involving that feature.
Additionally, disease nodes are defined as
$
\bm{\mathcal{V}}_{\text{disease}}
=
\left\{
v \in \bm{\mathcal{V}}
\;\middle|\;
\mathrm{Type}(v)=\text{Disease}
\right\}.
$
These memory indices enable subgraph extraction from microscopic features.
Literature retrieval is performed on PubMed using entity-augmented keyword queries with BM25 ranking (top-$K=50$ per query), and structured extraction is conducted by a pretrained GPT-4-class LLM ($T=0$) with a fixed prompt template.

\begin{figure*}[!h]
\centering
\includegraphics[width=1\textwidth]{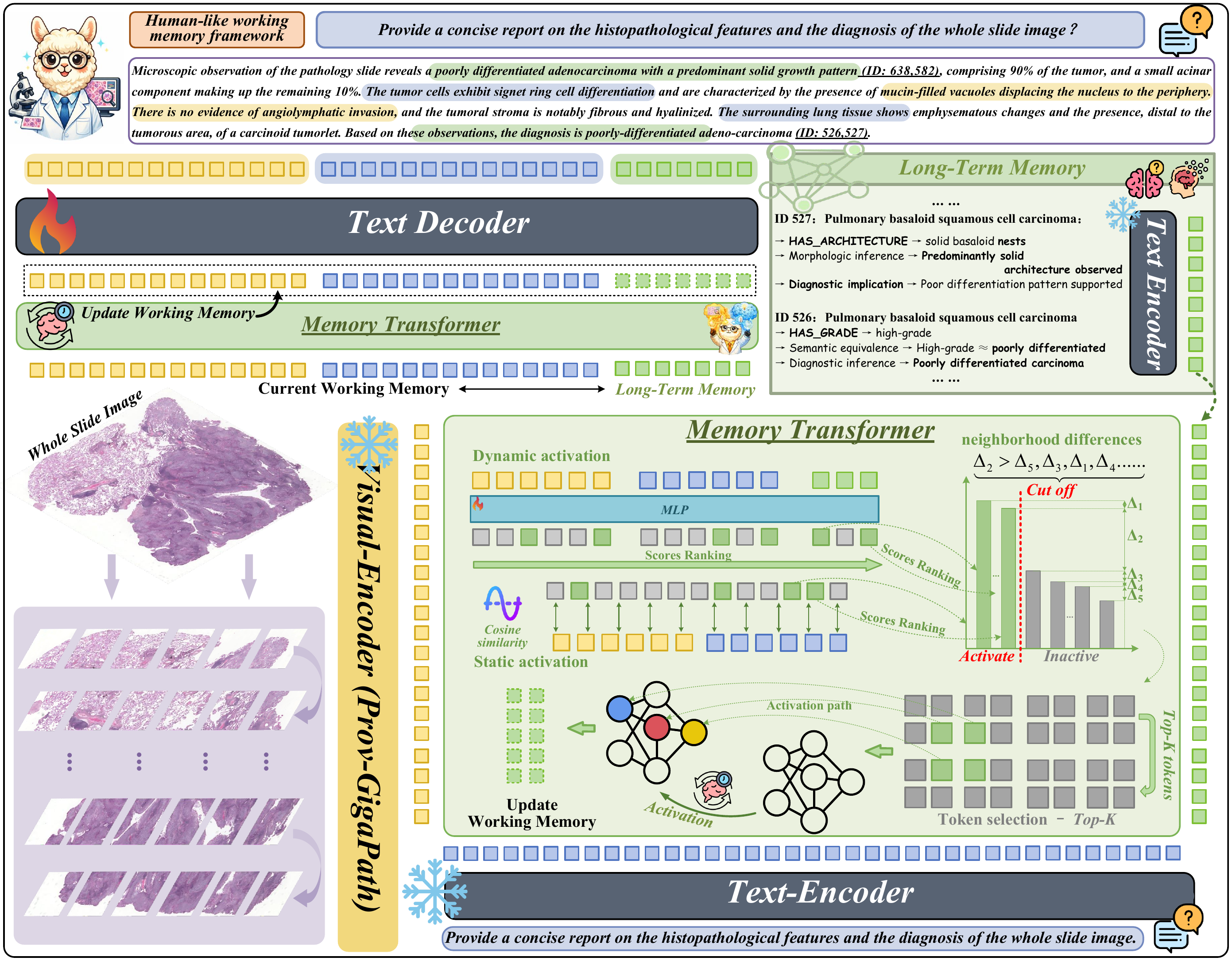}
\caption{Framework of \texttt{PathMem}.
A memory-augmented MLLMs for computational pathology that aligns visual, textual, and knowledge graph representations, and adaptively activates LTM for knowledge-grounded reasoning about pathology.}
\label{Figure2}
\end{figure*}

\subsection{Memory Transformer: Knowledge-Aware LTM$\rightarrow$WM Transition}
As illustrated in \textcolor{red}{Figure~\ref{Figure2}}, the \textbf{Memory Transformer} serves as the core module that bridges the LTM and WM through structured embedding interactions.
\textbf{\ding{168} LTM Embedding Space.}
Let the constructed knowledge graph define a structured LTM space
represented by an embedding bank
\begin{equation}
\bm{\mathcal{Q}}
=
\left\{
\mathbf{m}_i
\in
\mathbb{R}^{d}
\;\middle|\;
i=1,\dots,N
\right\},
\qquad
\bm{\mathcal{Q}}
\in
\mathbb{R}^{N \times d},
\end{equation}
here $N$ denotes the total number of knowledge entities stored in LTM,
$d$ is the shared embedding dimensionality with the multimodal backbone,
and each $\mathbf{m}_i$ encodes a graph-aware knowledge representation
learned from literature-grounded triples.
The memory space $\bm{\mathcal{Q}}$ is fixed during inference
and serves as an external, persistent knowledge reservoir.
Given an input multimodal sequence
$\mathbf{X} \in \mathbb{R}^{T \times d}$,
we compute a normalized query representation
$
\mathbf{q}
=
\bar{\mathbf{x}_t}
\left(
\|\bar{\mathbf{x}_t}\|_2 + \varepsilon
\right)^{-1},
$
here $\mathbf{x}_t$ denotes the $t$-th token embedding,
$\|\cdot\|_2$ is the Euclidean norm,
and $\varepsilon$ is a small constant for numerical stability.
The vector $\mathbf{q} \in \mathbb{R}^{d}$ summarizes the global semantic context of the current input.
\textbf{\ding{165} Structured Memory Selection.}
We define a relevance operator over the LTM space:
\begin{equation}
\bm{\mathcal{J}}
=
\mathrm{Softmax}
\left(
\frac{
\bm{\mathcal{Q}}
\mathbf{q}
}{
\sqrt{d}
}
+
\bm{\mathcal{N}}
\right),
\qquad
\bm{\mathcal{J}}
\in
\mathbb{R}^{N}.
\end{equation}
In this formulation,
$\bm{\mathcal{Q}}\mathbf{q}$ denotes the similarity between all memory tokens and the query,
$\frac{1}{\sqrt{d}}$ is the scaled dot-product normalization factor,
and $\bm{\mathcal{N}} \in \mathbb{R}^{N}$ is a structural masking vector
that suppresses previously selected indices or enforces diversity constraints,
we adopt a non-replacement top-$k$ masking strategy, where previously selected indices are assigned $-\infty$ in $\bm{\mathcal{N}}$ to prevent repeated selection and encourage diversity. The resulting vector $\bm{\mathcal{J}}$ represents a probability distribution over the entire memory bank,
encoding query-conditioned relevance scores.
To obtain a sparse WM,
we apply top-$k$ selection:
\begin{equation}
\bm{\mathcal{I}}
=
\operatorname{TopK}
\left(
\bm{\mathcal{J}},
k
\right),
\quad
\bm{\mathcal{Q}}_{\mathrm{work}}
=
\bm{\mathcal{J}}_{\bm{\mathcal{I}}}
\odot
\bm{\mathcal{Q}}_{\bm{\mathcal{I}}},
\end{equation}
here $k \ll N$ ensures sparsity,
$\bm{\mathcal{I}}$ denotes the selected indices,
$\bm{\mathcal{Q}}_{\bm{\mathcal{I}}} \in \mathbb{R}^{k \times d}$ is the retrieved subset,
and $\odot$ denotes row-wise scaling by relevance weights.
This operation converts dense LTM
into a compact WM block.
\textbf{\ding{171} Memory-Augmented Transformer Input.}
The WM is then prepended to the original sequence:
$
\mathbf{X}^{\star}
=
\mathrm{Concat}
\left(
\bm{\mathcal{Q}}_{\mathrm{work}},
\mathbf{X}
\right)
\in
\mathbb{R}^{(k+T) \times d}.
$
The augmented sequence $\mathbf{X}^{\star}$ is fed into the Transformer encoder,
allowing self-attention to jointly model
long-term knowledge tokens and input-specific representations.
Through this mechanism,
external structured memory is dynamically projected
into the model’s space,
enabling evidence-aware reasoning
without expanding parameters.
The model supports incremental memory updates without requiring full retraining of the Memory Transformer.

\section{Experiments}
\label{Experiments}
\vspace{-1.0em}

\textbf{\emph{Experimental Setup.}}
Experiments are primarily conducted on \textbf{WSI-Bench}, with additional zero-shot evaluation on external datasets to assess generalization. More details on data preprocessing are provided in \textcolor{red}{Appendix~\ref{appendix Dataset Processing}}.
\textbf{In-domain benchmark.}
We adopt \textbf{WSI-Bench} as the main benchmark for training and evaluation. Derived from TCGA, it contains 9,850 WSIs spanning 30 tumor types and 179,569 VQA pairs, covering morphology understanding, diagnosis, therapy reasoning, and report generation. We follow the official split, using 9,642 slides (175,450 QA pairs) for training and 208 slides (4,119 QA pairs) for testing. Evaluation follows the original protocol, including WSI-P, WSI-R, accuracy, BLEU, and ROUGE.
\textbf{Out-of-domain evaluation.}
We further assess generalization in a zero-shot manner on three external datasets: \textit{WSI-VQA}~\cite{chen2024wsi}, \textit{SlideBench-VQA (BCNB)}~\cite{chen2025slidechat}, and \textit{CPTAC-NSCLC}. These datasets are used strictly for evaluation without additional fine-tuning.
The details of \textbf{Long-term memory construction} are included in \textcolor{red}{Appendix~\ref{KG Construction}}.
\noindent\textbf{Implementation Details \& Metrics} are included in \textcolor{red}{Appendix~\ref{appendix Implementation Details}}.

\subsection{Main Result}

\noindent\textbf{Quantitative analysis}
\vspace{-1.0em}
\paragraph{\textbf{Obs. \ding{182}: Consistent Performance Gains of \texttt{PathMem} Across Pathological Tasks.}}
\textcolor{red}{Table \ref{tab_average_performance_11111}} presents the comparison on WSI-Bench. \texttt{PathMem} achieves the best overall average of 0.768, surpassing WSI-LLaVA at 0.754, Quilt LLaVA at 0.721, WSI-VQA at 0.590, and GPT-4o at 0.507, demonstrating superior comprehensive capability.
In Morphological Analysis, \texttt{PathMem} attains the highest open ended precision and close ended accuracy at 0.535 and 0.972, with competitive relevance at 0.542. For Diagnosis, it leads in precision, relevance, and accuracy with 0.707, 0.701, and 0.868, outperforming all baselines. In Treatment Planning, \texttt{PathMem} achieves 0.796 precision, 0.792 relevance, and 1.000 accuracy, maintaining the best balance across metrics.
Notably, thumbnail based models show clear degradation. GPT-4o yields only 0.220 precision and 0.471 accuracy in Morphological Analysis, highlighting the necessity of WSI level modeling. Overall, \texttt{PathMem} consistently improves open ended reasoning and achieves the strongest average performance across all pathological tasks.
\vspace{-1.0em}
\begin{table}[!h]
  \centering
    \setlength\tabcolsep{3.6pt}
    \caption{Quantitative evaluation of WSI and general-purpose MLLMs on our WSI-Bench across three pathological capabilities. WSI-P: WSI-Precision, WSI-R: WSI-Relevance, Acc: accuracy, open: open-ended question, and close: close-ended question.}
  \scalebox{0.7}{
  \begin{tabular}{l c c c c c c c c c c c}
    \Xhline{1.2pt}
    \multicolumn{2}{c}{\textbf{}} & \multicolumn{3}{c}{\textbf{Morphological Analysis}} & \multicolumn{3}{c}{\textbf{Diagnosis}} & \multicolumn{3}{c}{\textbf{Treatment Planning}} & \multirow{3}{*}{\textbf{Average}}\\
    \cmidrule(lr){3-5} \cmidrule(lr){6-8} \cmidrule(lr){9-11}
    Model & Input & \multicolumn{2}{c}{Open} & Close & \multicolumn{2}{c}{Open} & Close & \multicolumn{2}{c}{Open} & Close &\\
    & & WSI-P & WSI-R & Acc & WSI-P & WSI-R & Acc & WSI-P & WSI-R & Acc &\\
    \Xhline{1.2pt}
    Quilt-LLaVA \cite{seyfioglu2024quilt} & Thumbnail & 0.448 & 0.447 & 0.947& 0.586 & 0.604 &0.849& 0.788 & 0.816 & 1.000 & 0.721\\
    GPT-4o \cite{islam2024gpt} & Thumbnail & 0.220 & 0.204 &0.471& 0.472 & 0.457 &0.530& 0.496 & 0.841 & 0.875 & 0.507\\
    WSI-VQA \cite{chen2024wsi} & WSI & 0.395 & 0.462 &0.758& 0.436 & 0.525 & 0.577& 0.791 & 0.595 &0.771 &0.590\\
    WSI-LLaVA \cite{liang2025wsi} & WSI & 0.488 & 0.610 & 0.951 & 0.610 & 0.612 & 0.863 & \textbf{0.810} & \textbf{0.845} & 1.000 & 0.754\\
    \texttt{PathMem (Ours)} & WSI & \textcolor{blue}{0.535} & \textcolor{blue}{0.542} & \textcolor{blue}{0.972}& \textcolor{blue}{0.707} & \textcolor{blue}{0.701} & \textcolor{blue}{0.868}& 0.796 & 0.792 &\textcolor{blue}{1.000} &\textcolor{blue}{0.768}\\
    \Xhline{1.2pt}
  \end{tabular}
  }
  \label{tab_average_performance_11111}
\end{table}
\paragraph{\textbf{Obs. \ding{183}: Superior Report Generation Performance of \texttt{PathMem}.}}
\textcolor{red}{Table \ref{tab_report_gen22222}} reports the results for report generation. \texttt{PathMem} achieves the best performance across all metrics, with BLEU-1 to BLEU-4 scores of 0.548, 0.420, 0.347, and 0.302, consistently outperforming WSI-LLaVA (0.480, 0.358, 0.287, 0.240) and other baselines.
In semantic evaluation, \texttt{PathMem} attains 0.536 ROUGE-L and 0.531 METEOR, surpassing WSI-LLaVA (0.490 / 0.465) and significantly exceeding GPT-4o (0.132 / 0.167), indicating stronger content coverage and alignment with reference reports.
It also achieves the highest WSI-Precision and WSI-Relevance (0.508 / 0.530), compared with 0.380 / 0.429 for WSI-LLaVA, reflecting improved slide-level grounding and clinical relevance.
Overall, consistent gains across lexical, semantic, and WSI-specific metrics demonstrate the superiority of \texttt{PathMem} in generating accurate and pathology-coherent reports.
\begin{table}[!h]
    \setlength\tabcolsep{3.9pt}
    \centering
    \caption{Quantitative evaluation of WSI and general MLLMs in report generation on WSI-Bench.}
    \scalebox{0.64}{
    \begin{tabular}{lcccccccc}
        \Xhline{1.2pt}
        Models & BLEU-1 & BLEU-2 & BLEU-3 & BLEU-4 & ROUGE-L & METEOR & WSI-Precision & WSI-Relevance \\
        \Xhline{1.2pt}
        GPT-4o \cite{islam2024gpt} &  0.202&0.069 &0.030 &0.016 &0.132 & 0.167& 0.067 &0.138 \\
        Quilt-LLaVA \cite{seyfioglu2024quilt} & 0.474 & 0.351 & 0.282 & 0.236 & 0.475 & 0.460 & 0.324 & 0.333 \\
        MI-Gen \cite{chen2024wsicaption} & 0.403 & 0.306 & 0.248 & 0.209 & 0.446 & 0.407 & 0.310 & 0.377 \\
        Hist-Gen \cite{guo2024histgen} & 0.406 & 0.307 & 0.248& 0.208 & 0.448 & 0.416 & 0.300 & 0.397 \\
        WSI-LLaVA \cite{liang2025wsi}& 0.480 & 0.358 & 0.287 & 0.240 & 0.490 & 0.465 &0.380 & 0.429\\
        \texttt{PathMem (Ours)} & \textcolor{blue}{0.548} \color{blue}$\pm$ \color{blue}0.151 & \textcolor{blue}{0.420} \color{blue}$\pm$ \color{blue}0.173 & \textcolor{blue}{0.347} \color{blue}$\pm$ \color{blue}0.188 & \textcolor{blue}{0.302} \color{blue}$\pm$ \color{blue}0.196 & \textcolor{blue}{0.536} \color{blue}$\pm$ \color{blue}0.155 & \textcolor{blue}{0.531} \color{blue}$\pm$ \color{blue}0.158 & \textcolor{blue}{0.508} \color{blue}$\pm$ \color{blue}0.147 & \textcolor{blue}{0.530} \color{blue}$\pm$ \color{blue}0.144 \\
        \Xhline{1.2pt}
        \vspace{-3.0em}
    \end{tabular}
    }
    \label{tab_report_gen22222}
\end{table}

\paragraph{\textbf{Obs. \ding{184}: Strong Zero-Shot Generalization of \texttt{PathMem} Across External Benchmarks.}}
\textcolor{red}{Table \ref{tab_external_validation_33333}} presents the zero-shot external validation results. \texttt{PathMem} achieves the best performance on all benchmarks, demonstrating strong generalization across datasets and tasks.
On WSI-VQA, \texttt{PathMem} attains 0.572, outperforming WSI-LLaVA at 0.546 and substantially exceeding Quilt-LLaVA at 0.132. On SlideBench VQA BCNB, \texttt{PathMem} achieves the highest scores in all categories, including Tumor at 0.918, Grade at 0.482, and Subtype at 0.313, yielding the best average of 0.571. In comparison, WSI-LLaVA obtains an average of 0.553, while WSI-VQA shows clear instability with an average of 0.113 and zero performance on Subtype.
On CPTAC NSCLC, \texttt{PathMem} reaches 0.754, surpassing WSI-LLaVA at 0.721 and markedly improving over Quilt-LLaVA at 0.603. Overall, consistent gains across external datasets confirm the robustness and cross domain generalization ability of \texttt{PathMem}.

\begin{table}[!h]
\centering
\caption{Zero-shot external validation on WSI-VQA, SlideBench-VQA, and CPTAC-NSCLC.}
\small
\setlength{\tabcolsep}{4pt}
\resizebox{0.7\linewidth}{!}{
\begin{tabular}{l c c c c c c c}
\Xhline{1.2pt}
\multirow{2}{*}{Model} & \multirow{2}{*}{Input} & \multirow{2}{*}{\textbf{WSI-VQA}}
& \multicolumn{4}{c}{\textbf{SlideBench-VQA (BCNB)}}
& \multirow{2}{*}{\textbf{CPTAC}} \\
\cmidrule(lr){4-7}
& & & Tumor & Grade & Subtype & Avg & \\
\Xhline{1.2pt}
Quilt-LLaVA\cite{seyfioglu2024quilt} & Thumb & 0.132 & 0.113 & 0.194 & 0.181 & 0.163 & 0.603 \\
WSI-VQA\cite{chen2024wsi} & WSI & 0.469 & 0.039 & 0.300 & 0.000 & 0.113 & 0.000 \\
SlideChat\cite{chen2025slidechat} & WSI & - & 0.902 & 0.231 & 0.175 & 0.436 & - \\
WSI-LLaVA\cite{liang2025wsi} & WSI & 0.546 & 0.905 & 0.463 & 0.292 & 0.553 & 0.721 \\
\texttt{PathMem (Ours)} & WSI & \textbf{0.572} & \textbf{0.918} & \textbf{0.482} & \textbf{0.313} & \textbf{0.571} & \textbf{0.754} \\
\Xhline{1.2pt}
\end{tabular}
}
\label{tab_external_validation_33333}
\end{table}

\begin{figure*}[h]
\centering
\includegraphics[width=0.95\textwidth]{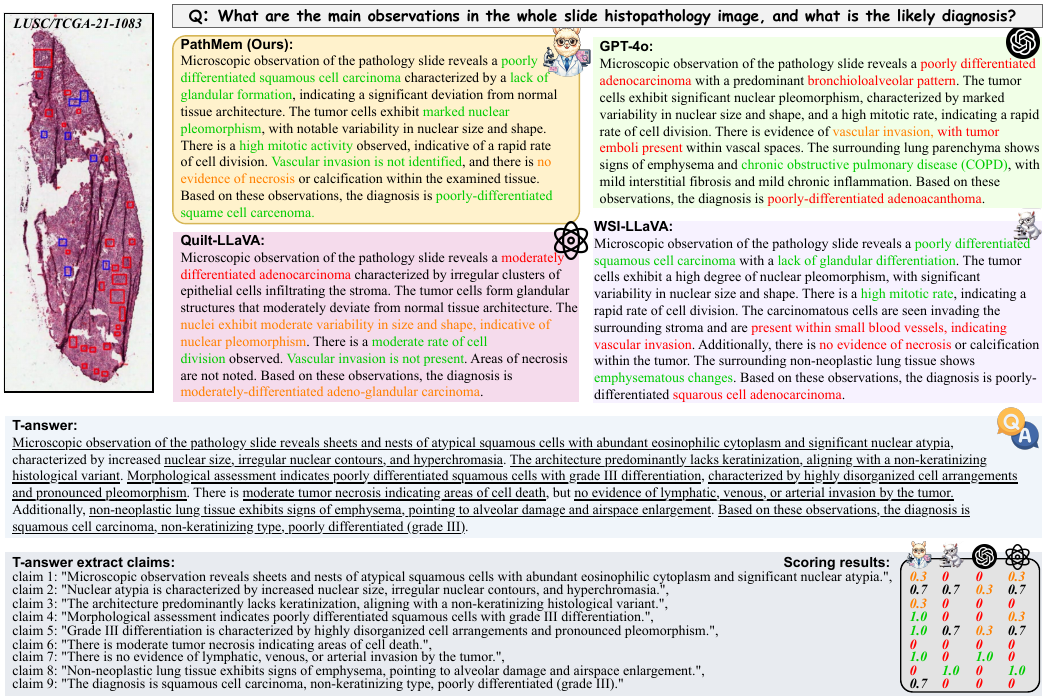}
\caption{Qualitative comparison of generated reports by our method and three baseline approaches on the report generation task.
(Red highlights denote incorrect content, while green highlights denote correct content, and orange highlights denote missing content from T-answer.)
}
\label{444444444444444444444}
\end{figure*}

\noindent\textbf{Qualitative analysis}

\paragraph{\textbf{Obs. \ding{182}: The Superior Reasoning Capability of \texttt{PathMem}.}}
As shown in \textcolor{red}{Figure \ref{444444444444444444444}}, we present a qualitative comparison for case \textbf{LUSC/TCGA-21-1083}. The ground truth is \textit{poorly differentiated (Grade III), non-keratinizing squamous cell carcinoma}.
\textbf{Overall Diagnosis.}
\texttt{PathMem} correctly predicts poorly differentiated squamous cell carcinoma, fully consistent with the reference. WSI-LLaVA captures squamous features but outputs an inconsistent hybrid diagnosis. GPT-4o and Quilt-LLaVA misclassify the case as adenocarcinoma variants, reflecting confusion between squamous and glandular differentiation.
\textbf{Fine-Grained Morphology.}
For squamous differentiation (Claims 1, 3–5), \texttt{PathMem} best captures poor differentiation and marked pleomorphism, with partial recognition of non-keratinizing features. Other models show weaker alignment and often hallucinate glandular structures.
For nuclear atypia (Claim 2), all models detect pleomorphism, but \texttt{PathMem} demonstrates greater consistency.
For necrosis and invasion (Claims 6–7), performance is generally limited; however, \texttt{PathMem} correctly identifies the absence of vascular invasion.
For background lung changes (Claim 8), WSI-LLaVA and Quilt-LLaVA note emphysematous features, whereas \texttt{PathMem} focuses on tumor characteristics.
\paragraph{\textbf{Obs. \ding{183}: Metric Evidence Supporting Reasoning Strength of \texttt{PathMem}.}}
The quantitative results closely mirror the color-coded qualitative comparison (\textcolor{red}{\textit{Appendix Figure} \ref{KG-activationV2}}).
\textbf{Hist-Gen} exhibits numerous red deviations and orange omissions, leading to low semantic overlap (BLEU-4: 0.025; ROUGE-L: 0.182) and limited pathological relevance (WSI-P: 0.257; WSI-R: 0.230).
\textbf{WSI-LLaVA} shows moderate improvement, with fewer deviations but continued omission of key invasive features. This corresponds to slightly higher BLEU-4 (0.043), ROUGE-L (0.250), and WSI-based metrics (WSI-P: 0.311; WSI-R: 0.294).
In contrast, \texttt{PathMem} demonstrates predominantly green matches with minimal deviations, alongside multiple blue knowledge graph–grounded concepts. This results in substantially improved lexical and clinical alignment (BLEU-4: 0.361; ROUGE-L: 0.583; METEOR: 0.678) and markedly higher WSI-Precision and recall (0.596 / 0.592).
Overall, reductions in deviations and missing findings correspond directly to gains across both NLU and WSI-based metrics, while knowledge graph grounding further enhances diagnostic consistency.

\noindent
\textbf{\ding{168} Answer-Aligned Memory Path.}
Visual Cues (solid growth, poor differentiation, papillary areas, vascular invasion, fibrotic lung background)
$\;\rightarrow\;$
\textbf{Static LTM Retrieval} (lung primary site, NSCLC spectrum, solid/high-grade architecture)
$\;\rightarrow\;$
\textbf{Dynamic Multimodal Reweighting}
($\uparrow$ glandular signals + $\uparrow$ squamous patterns; $\downarrow$ irrelevant entities)
$\;\rightarrow\;$
\textbf{Adaptive LTM$\rightarrow$WM Transfer}
(lung context + dual-lineage semantics + invasion knowledge)
$\;\rightarrow\;$
\textbf{WM-Integrated Diagnosis:}
\textbf{\emph{Poorly Differentiated Adenosquamous Carcinoma}}
(+ \textbf{angiolymphatic invasion} [\textit{LTM}]
+ \textbf{focal visceral pleural involvement} [\textit{LTM}]
+ \textbf{background emphysema} [\textit{LTM}]).
The highlighted diagnostic elements emerge through structured LTM activation and controlled LTM$\rightarrow$WM transition, contributing to the final textual output beyond purely parametric generation.
\begin{table}[!h]
\centering
\caption{Ablation study on static and dynamic KG retrieval. The full model with both static and dynamic LTM achieves the best performance across all metrics. \textbf{D denotes Dynamic-activation, and S denotes Static-activation LTM}.}
{ \renewcommand{\arraystretch}{1.3}
\resizebox{1\textwidth}{!}{
\begin{tabular}{ c c c c c | c|  c|  c|  c|  c|  c|  c}
\Xhline{1.2pt}
\multirow{2}{*}{Version} & \cellcolor{CadetBlue!20}Baseline & \multicolumn{2}{c}{\cellcolor{CadetBlue!20} LTM} & \multicolumn{8}{c}{\cellcolor{CadetBlue!20}\texttt{PathMem}} \\
&\cellcolor{CadetBlue!20} -
&\cellcolor{CadetBlue!20} D &\cellcolor{CadetBlue!20} S

&\cellcolor{CadetBlue!20} BLEU-1
&\cellcolor{CadetBlue!20} BLEU-2
&\cellcolor{CadetBlue!20} BLEU-3
&\cellcolor{CadetBlue!20} BLEU-4
&\cellcolor{CadetBlue!20} ROUGE-L
&\cellcolor{CadetBlue!20} METEOR
&\cellcolor{CadetBlue!20} WSI-P
&\cellcolor{CadetBlue!20} WSI-R \\
\Xhline{1.2pt}

a & \checkmark & & &
\cellcolor{gray!10}0.481 $\pm$ 0.181  &
\cellcolor{gray!10}0.357 $\pm$ 0.196  &
\cellcolor{gray!10}0.287 $\pm$ 0.205  &
\cellcolor{gray!10}0.241 $\pm$ 0.212  &
\cellcolor{gray!10}0.477 $\pm$ 0.177  &
\cellcolor{gray!10}0.464 $\pm$ 0.178  &
\cellcolor{gray!10}0.404 $\pm$ 0.162  &
\cellcolor{gray!10}0.445 $\pm$ 0.163  \\
b & \checkmark & \checkmark & &
\cellcolor{gray!1}0.528 $\pm$ 0.153 &
\cellcolor{gray!1}0.405 $\pm$ 0.176 &
\cellcolor{gray!1}0.332 $\pm$ 0.195 &
\cellcolor{gray!1}0.284 $\pm$ 0.198 &
\cellcolor{gray!1}0.518 $\pm$ 0.168 &
\cellcolor{gray!1}0.508 $\pm$ 0.164 &
\cellcolor{gray!1}0.492 $\pm$ 0.155 &
\cellcolor{gray!1}0.516 $\pm$ 0.147
 \\
c & \checkmark & & \checkmark &
\cellcolor{gray!10}0.521 $\pm$ 0.161   &
\cellcolor{gray!10}0.396 $\pm$ 0.183  &
\cellcolor{gray!10}0.324 $\pm$ 0.205  &
\cellcolor{gray!10}0.278 $\pm$ 0.205  &
\cellcolor{gray!10}0.510 $\pm$ 0.170  &
\cellcolor{gray!10}0.511 $\pm$ 0.163  &
\cellcolor{gray!10}0.489 $\pm$ 0.158  &
\cellcolor{gray!10}0.510 $\pm$ 0.148  \\
d & \checkmark & \checkmark & \checkmark &
\cellcolor{yellow!20}\textbf{0.548} $\pm$ 0.151 &
\cellcolor{yellow!20}\textbf{0.420} $\pm$ 0.173 &
\cellcolor{yellow!20}\textbf{0.347} $\pm$ 0.188 &
\cellcolor{yellow!20}\textbf{0.302} $\pm$ 0.196 &
\cellcolor{yellow!20}\textbf{0.536} $\pm$ 0.155 &
\cellcolor{yellow!20}\textbf{0.531} $\pm$ 0.158 &
\cellcolor{yellow!20}\textbf{0.508} $\pm$ 0.147 &
\cellcolor{yellow!20}\textbf{0.530} $\pm$ 0.144 \\
\Xhline{1.2pt}
\vspace{-3.5em}
\end{tabular}}
}
\label{Table3}
\end{table}
\subsection{Analysis and Discussion}
\paragraph{\textbf{Obs. \ding{182}: Ablation Study on Static and Dynamic KG Retrieval.}}
\textcolor{red}{Table \ref{Table3}} presents the ablation results on static and dynamic KG retrieval. Both mechanisms consistently improve performance over the baseline (Version a). Adding dynamic LTM (Version b) increases BLEU-4 from 0.241 to 0.284 and WSI-P from 0.404 to 0.492. Static LTM (Version c) also yields clear gains, with BLEU-4 reaching 0.278 and WSI-R 0.510.
Dynamic LTM achieves slightly higher BLEU scores than static LTM, for example 0.284 versus 0.278 in BLEU-4, while static LTM attains a marginally higher METEOR score, 0.511 versus 0.508.
The full model (Version d) achieves the best results on all metrics, including 0.302 BLEU-4, 0.536 ROUGE-L, 0.531 METEOR, 0.508 WSI-P, and 0.530 WSI-R. Compared with the baseline, BLEU-4 improves by 0.061 and WSI-P by 0.104.
These results indicate that static and dynamic retrieval provide complementary benefits, and their combination is essential for optimal performance.
\emph{\textbf{Attribution analysis of performance improvement}}
Specifically, Version b \& c primarily reflect the benefit of incorporating external knowledge through dynamic or static retrieval mechanisms, respectively. In this sense, their improvements over the baseline can be largely attributed to the introduction of high-quality pathology knowledge.
In contrast, Version d introduces the full memory controller that jointly orchestrates dynamic and static knowledge activation. Therefore, the additional gains observed from Version d over Versions b and c can be more directly attributed to the effectiveness of the proposed memory control mechanism.
\textit{\textbf{This suggests that the memory controller plays a critical role in coordinating heterogeneous knowledge sources and enabling their complementary benefits, beyond what can be achieved by naive retrieval alone}}.
\vspace{-1.0em}
\begin{table}[!h]
\centering
\caption{Sensitivity analysis on the maximum number of activated KG tokens (Top-K). Performance comparison under different limits for dynamic and static LTM retrieval. \textbf{D denotes Dynamic-activation, and S denotes Static-activation LTM}.}
{ \renewcommand{\arraystretch}{1.3}
\resizebox{1\textwidth}{!}{
\begin{tabular}{ c c c c | c | c | c | c | c | c | c}
\Xhline{1.2pt}
\multirow{2}{*}{Version} & \multicolumn{2}{c}{\cellcolor{CadetBlue!20} Token Cap} & \multicolumn{8}{c}{\cellcolor{CadetBlue!20}\texttt{PathMem}} \\
&\cellcolor{CadetBlue!20} D &\cellcolor{CadetBlue!20} S
&\cellcolor{CadetBlue!20} BLEU-1
&\cellcolor{CadetBlue!20} BLEU-2
&\cellcolor{CadetBlue!20} BLEU-3
&\cellcolor{CadetBlue!20} BLEU-4
&\cellcolor{CadetBlue!20} ROUGE-L
&\cellcolor{CadetBlue!20} METEOR
&\cellcolor{CadetBlue!20} WSI-P
&\cellcolor{CadetBlue!20} WSI-R \\
\Xhline{1.2pt}

a & 1 & 1 &
\cellcolor{gray!10}0.517 $\pm$ 0.161  &
\cellcolor{gray!10}0.393 $\pm$ 0.184  &
\cellcolor{gray!10}0.320 $\pm$ 0.198  &
\cellcolor{gray!10}0.274 $\pm$ 0.210  &
\cellcolor{gray!10}0.508 $\pm$ 0.172  &
\cellcolor{gray!10}0.500 $\pm$ 0.169  &
\cellcolor{gray!10}0.477 $\pm$ 0.163  &
\cellcolor{gray!10}0.496 $\pm$ 0.151  \\
b & 2 & 2 &
\cellcolor{gray!1}0.525 $\pm$ 0.160 &
\cellcolor{gray!1}0.402 $\pm$ 0.181 &
\cellcolor{gray!1}0.330 $\pm$ 0.204 &
\cellcolor{gray!1}0.284 $\pm$ 0.205 &
\cellcolor{gray!1}0.517 $\pm$ 0.168 &
\cellcolor{gray!1}0.509 $\pm$ 0.164 &
\cellcolor{gray!1}0.484 $\pm$ 0.159 &
\cellcolor{gray!1}0.497 $\pm$ 0.148 \\
c & 3 & 3 &
\cellcolor{gray!10}0.531 $\pm$ 0.153  &
\cellcolor{gray!10}0.410 $\pm$ 0.175  &
\cellcolor{gray!10}0.339 $\pm$ 0.193  &
\cellcolor{gray!10}0.293 $\pm$ 0.196  &
\cellcolor{gray!10}0.525 $\pm$ 0.162  &
\cellcolor{gray!10}0.514 $\pm$ 0.161  &
\cellcolor{gray!10}0.493 $\pm$ 0.155  &
\cellcolor{gray!10}0.513 $\pm$ 0.147  \\
d & 4 & 4 &
\cellcolor{gray!1}0.537 $\pm$ 0.154 &
\cellcolor{gray!1}0.414 $\pm$ 0.173 &
\cellcolor{gray!1}0.340 $\pm$ 0.194 &
\cellcolor{gray!1}0.292 $\pm$ 0.197 &
\cellcolor{gray!1}0.528 $\pm$ 0.158 &
\cellcolor{gray!1}0.517 $\pm$ 0.161 &
\cellcolor{gray!1}0.502 $\pm$ 0.149 &
\cellcolor{gray!1}0.526 $\pm$ 0.146 \\
g & 5 & 5 &
\cellcolor{yellow!20}\textbf{0.542} $\pm$ 0.151 &
\cellcolor{yellow!20}\textbf{0.418} $\pm$ 0.174 &
\cellcolor{yellow!20}\textbf{0.346} $\pm$ 0.188 &
\cellcolor{yellow!20}\textbf{0.298} $\pm$ 0.197 &
\cellcolor{yellow!20}\textbf{0.532} $\pm$ 0.156 &
\cellcolor{yellow!20}\textbf{0.523} $\pm$ 0.158 &
\cellcolor{yellow!20}\textbf{0.506} $\pm$ 0.147 &
\cellcolor{yellow!20}\textbf{0.528} $\pm$ 0.143 \\
\Xhline{1.2pt}
\vspace{-2.5em}
\end{tabular}}
}
\label{Table5555555555555555}
\end{table}

\paragraph{\textbf{Obs. \ding{183}: Sensitivity Analysis of Top-K KG Tokens.}}
\textcolor{red}{Table \ref{Table5555555555555555}} presents the sensitivity analysis on the maximum number of activated KG tokens for dynamic and static LTM retrieval. Performance consistently improves as Top-K increases from 1 to 5.
With Top-K equals 1, BLEU-4 is 0.274 and WSI-P is 0.477. Increasing Top-K to 3 raises BLEU-4 to 0.293 and WSI-R to 0.513. The best results are achieved at Top-K equals 5, with 0.298 BLEU-4, 0.532 ROUGE-L, 0.523 METEOR, 0.506 WSI-P, and 0.528 WSI-R. Compared with Top-K equals 1, BLEU-4 improves by 0.024 and WSI-P by 0.029.
The gains diminish beyond Top-K equals 3, for example BLEU-4 increases only from 0.293 to 0.298 when moving from 3 to 5. This indicates that a moderate token budget captures most useful knowledge, while larger limits provide limited but consistent improvements.
\paragraph{\textbf{Obs. \ding{184}: Inference Efficiency Analysis}.}

\begin{wraptable}{l}{0.61\linewidth}
\vspace{-12pt}
\centering
\small
\caption{Efficiency and practicality analysis of \texttt{PathMem}.}
\label{tab_pathmem_efficiency666666}
\begin{tabular}{lc}
\Xhline{1.0pt}
\cellcolor{CadetBlue!20}Metric & \cellcolor{CadetBlue!20}Value \\
\Xhline{1.0pt}
Inference Time & $\sim$1.45 items (WSI) / s
\\
\cellcolor{gray!10}GPU Utilization & \cellcolor{gray!10}1 $\times$ GPU  \\
GPU Memory Usage & 16.3 GB   \\
\cellcolor{gray!10}Image Feature Size & \cellcolor{gray!10}$1728 \times 4096$ tokens  \\
KG Construction Cost & Offline embedding + indexing  \\
\cellcolor{gray!10}KG Update Cost & \cellcolor{gray!10}Re-embedding + re-indexing  \\
\Xhline{1.0pt}
\end{tabular}
\vspace{-6pt}
\end{wraptable}
As listed in \textcolor{red}{Table \ref{tab_pathmem_efficiency666666}}, \texttt{PathMem} runs at $\sim$1.45 items/s with 16.3 GB memory on a single GPU, indicating moderate inference overhead with good scalability.
The additional cost mainly comes from KG construction and updates (embedding and indexing), which are performed offline and amortized over repeated inference.
Overall, \texttt{PathMem} provides a practical efficiency--performance trade-off for high-quality reasoning tasks.
Considering that computational pathology workflows are typically performed at slide-level or case-level rather than under strict real-time constraints, \textbf{\textit{the current inference efficiency is sufficient for real-world clinical pathology deployment and routine diagnostic workflows.}}

\vspace{-0.5em}
\section{Conclusion \& Limitation}
\label{sec:Conclusion}
\vspace{-0.5em}
We propose \texttt{PathMem}, a memory-augmented MLLM for computational pathology that explicitly models long- and working-memory through a PubMed-derived pathology knowledge graph and a dual-stage Memory Transformer. Experiments on WSI-Bench and three zero-shot benchmarks show consistent improvements over SOTA models across core pathology tasks. Ablation and sensitivity studies confirm the effectiveness of dual-mode KG retrieval and moderate Top-K token settings.
\textbf{Limitation.}
\texttt{PathMem} introduces additional design complexity, which may require careful tuning of hyperparameters across tasks. The current implementation focuses on selected pathology benchmarks, and extending evaluation to a wider range of datasets could further assess generality. Moreover, the framework assumes a fixed knowledge retrieval strategy, leaving room for exploring more adaptive or task-specific retrieval mechanisms in future work.

\bibliography{references}

@String(AAAI = {AAAI})

@inproceedings{liang2025wsi,
  title={Wsi-llava: A multimodal large language model for whole slide image},
  author={Liang, Yuci and Lyu, Xinheng and Chen, Wenting and Ding, Meidan and Zhang, Jipeng and He, Xiangjian and Wu, Song and Xing, Xiaohan and Yang, Sen and Wang, Xiyue and others},
  booktitle={Proceedings of the IEEE/CVF International Conference on Computer Vision},
  pages={22718--22727},
  year={2025}
}

@inproceedings{seyfioglu2024quilt,
  title={Quilt-LLaVA: Visual Instruction Tuning by Extracting Localized Narratives from Open-Source Histopathology Videos},
  author={Seyfioglu, Mehmet Saygin and Ikezogwo, Wisdom O and Ghezloo, Fatemeh and Krishna, Ranjay and Shapiro, Linda},
  booktitle={Proceedings of the IEEE/CVF Conference on Computer Vision and Pattern Recognition},
  pages={13183--13192},
  year={2024}
}

@inproceedings{chen2024wsicaption,
  title={WsiCaption: Multiple Instance Generation of Pathology Reports for Gigapixel Whole-Slide Images},
  author={Chen, Pingyi and Li, Honglin and Zhu, Chenglu and Zheng, Sunyi and Shui, Zhongyi and Yang, Lin},
  booktitle={International Conference on Medical Image Computing and Computer-Assisted Intervention},
  pages={546--556},
  year={2024},
  organization={Springer}
}

@inproceedings{guo2024histgen,
  title={Histgen: Histopathology report generation via local-global feature encoding and cross-modal context interaction},
  author={Guo, Zhengrui and Ma, Jiabo and Xu, Yingxue and Wang, Yihui and Wang, Liansheng and Chen, Hao},
  booktitle={International Conference on Medical Image Computing and Computer-Assisted Intervention},
  pages={189--199},
  year={2024},
  organization={Springer}
}

@article{islam2024gpt,
  title={Gpt-4o: The cutting-edge advancement in multimodal llm},
  author={Islam, Raisa and Moushi, Owana Marzia},
  journal={Authorea Preprints},
  year={2024},
  publisher={Authorea}
}

@article{shaikovski2024prism,
  title={PRISM: A Multi-Modal Generative Foundation Model for Slide-Level Histopathology},
  author={Shaikovski, George and Casson, Adam and Severson, Kristen and Zimmermann, Eric and Wang, Yi Kan and Kunz, Jeremy D and Retamero, Juan A and Oakley, Gerard and Klimstra, David and Kanan, Christopher and others},
  journal={arXiv preprint arXiv:2405.10254},
  year={2024}
}

@article{xu2024whole,
  title={A whole-slide foundation model for digital pathology from real-world data},
  author={Xu, Hanwen and Usuyama, Naoto and Bagga, Jaspreet and Zhang, Sheng and Rao, Rajesh and Naumann, Tristan and Wong, Cliff and Gero, Zelalem and Gonz{\'a}lez, Javier and Gu, Yu and others},
  journal={Nature},
  pages={1--8},
  year={2024},
  publisher={Nature Publishing Group UK London}
}

@inproceedings{sun2025cpath,
  title={Cpath-omni: A unified multimodal foundation model for patch and whole slide image analysis in computational pathology},
  author={Sun, Yuxuan and Si, Yixuan and Zhu, Chenglu and Gong, Xuan and Zhang, Kai and Chen, Pingyi and Zhang, Ye and Shui, Zhongyi and Lin, Tao and Yang, Lin},
  booktitle={Proceedings of the Computer Vision and Pattern Recognition Conference},
  pages={10360--10371},
  year={2025}
}

@article{ahmed2024pathalign,
  title={Pathalign: A vision-language model for whole slide images in histopathology},
  author={Ahmed, Faruk and Sellergren, Andrew and Yang, Lin and Xu, Shawn and Babenko, Boris and Ward, Abbi and Olson, Niels and Mohtashamian, Arash and Matias, Yossi and Corrado, Greg S and others},
  journal={arXiv preprint arXiv:2406.19578},
  year={2024}
}

@inproceedings{chen2025slidechat,
  title={Slidechat: A large vision-language assistant for whole-slide pathology image understanding},
  author={Chen, Ying and Wang, Guoan and Ji, Yuanfeng and Li, Yanjun and Ye, Jin and Li, Tianbin and Hu, Ming and Yu, Rongshan and Qiao, Yu and He, Junjun},
  booktitle={Proceedings of the Computer Vision and Pattern Recognition Conference},
  pages={5134--5143},
  year={2025}
}

@article{kim2025chatexaonepath,
  title={ChatEXAONEPath: An Expert-level Multimodal Large Language Model for Histopathology Using Whole Slide Images},
  author={Kim, Sangwook and Lee, Soonyoung and Jang, Jongseong},
  journal={arXiv preprint arXiv:2504.13023},
  year={2025}
}

@article{ding2025multimodal,
  title={A multimodal whole-slide foundation model for pathology},
  author={Ding, Tong and Wagner, Sophia J and Song, Andrew H and Chen, Richard J and Lu, Ming Y and Zhang, Andrew and Vaidya, Anurag J and Jaume, Guillaume and Shaban, Muhammad and Kim, Ahrong and others},
  journal={Nature medicine},
  pages={1--13},
  year={2025},
  publisher={Nature Publishing Group US New York}
}

@article{hu2025loc,
  title={LoC-Path: Learning to Compress for Pathology Multimodal Large Language Models},
  author={Hu, Qingqiao and Lyu, Weimin and Xu, Meilong and Qi, Kehan and Hu, Xiaoling and Gupta, Saumya and Zhou, Jiawei and Chen, Chao},
  journal={arXiv preprint arXiv:2512.05391},
  year={2025}
}

@inproceedings{sun2024pathasst,
  title={Pathasst: A generative foundation ai assistant towards artificial general intelligence of pathology},
  author={Sun, Yuxuan and Zhu, Chenglu and Zheng, Sunyi and Zhang, Kai and Sun, Lin and Shui, Zhongyi and Zhang, Yunlong and Li, Honglin and Yang, Lin},
  booktitle={Proceedings of the AAAI Conference on Artificial Intelligence},
  volume={38},
  number={5},
  pages={5034--5042},
  year={2024}
}

@article{zhang2025patho,
  title={Patho-AgenticRAG: towards multimodal agentic retrieval-augmented generation for pathology VLMs via reinforcement learning},
  author={Zhang, Wenchuan and Guo, Jingru and Zhang, Hengzhe and Zhang, Penghao and Chen, Jie and Zhang, Shuwan and Zhang, Zhang and Yi, Yuhao and Bu, Hong},
  journal={arXiv preprint arXiv:2508.02258},
  year={2025}
}

@article{ma2025generalizable,
  title={A generalizable pathology foundation model using a unified knowledge distillation pretraining framework},
  author={Ma, Jiabo and Guo, Zhengrui and Zhou, Fengtao and Wang, Yihui and Xu, Yingxue and Li, Jinbang and Yan, Fang and Cai, Yu and Zhu, Zhengjie and Jin, Cheng and others},
  journal={Nature Biomedical Engineering},
  pages={1--20},
  year={2025},
  publisher={Nature Publishing Group UK London}
}

@article{smart1995model,
  title={A model for medical knowledge representation application to the analysis of descriptive pathology reports},
  author={Smart, JF and Roux, M},
  journal={Methods of information in medicine},
  volume={34},
  number={04},
  pages={352--360},
  year={1995},
  publisher={Schattauer GmbH}
}

@article{huang2025knowledge,
  title={Knowledge-guided adaptation of pathology foundation models effectively improves cross-domain generalization and demographic fairness},
  author={Huang, Yanyan and Zhao, Weiqin and Zhang, Zhengyu and Chen, Yihang and Fu, Yu and Wu, Feng and Jiang, Yuming and Liang, Li and Wang, Shujun and Yu, Lequan},
  journal={Nature Communications},
  year={2025},
  publisher={Nature Publishing Group UK London}
}

@article{jiang2026pathreasoner,
  title={PathReasoner-R1: Instilling Structured Reasoning into Pathology Vision-Language Model via Knowledge-Guided Policy Optimization},
  author={Jiang, Songhan and Liu, Fengchun and Wang, Ziyue and Cai, Linghan and Zhang, Yongbing},
  journal={arXiv preprint arXiv:2601.21617},
  year={2026}
}

@article{zhao2024aligning,
  title={Aligning knowledge concepts to whole slide images for precise histopathology image analysis},
  author={Zhao, Weiqin and Guo, Ziyu and Fan, Yinshuang and Jiang, Yuming and Yeung, Maximus CF and Yu, Lequan},
  journal={npj Digital Medicine},
  volume={7},
  number={1},
  pages={383},
  year={2024},
  publisher={Nature Publishing Group UK London}
}

@article{he2025prior,
  title={Prior knowledge-guided multimodal deep learning system for biomarker exploration and prognosis prediction of urothelial carcinoma},
  author={He, Quanhao and Tan, Hao and Xiao, Bangxin and Tan, Yiwen and Peng, Xiang and Peng, Canjie and Yue, Xiaofeng and Jiang, Linshan and Cao, Youde and Lv, Fa Jin and others},
  journal={npj Digital Medicine},
  year={2025},
  publisher={Nature Publishing Group UK London}
}

@article{shi2025kpvg,
  title={KPVG: Knowledge-prompted vision-genomics model for cancer survival prediction in whole slide images},
  author={Shi, Jiangbo and Li, Chen and Zheng, Yefeng and Fu, Huazhu},
  journal={Information Fusion},
  pages={103660},
  year={2025},
  publisher={Elsevier}
}

@inproceedings{chen2024wsi,
  title={Wsi-vqa: Interpreting whole slide images by generative visual question answering},
  author={Chen, Pingyi and Zhu, Chenglu and Zheng, Sunyi and Li, Honglin and Yang, Lin},
  booktitle={European Conference on Computer Vision},
  pages={401--417},
  year={2024},
  organization={Springer}
}

@article{omar2024chatgpt,
  title={ChatGPT for digital pathology research},
  author={Omar, Mohamed and Ullanat, Varun and Loda, Massimo and Marchionni, Luigi and Umeton, Renato},
  journal={The Lancet Digital Health},
  volume={6},
  number={8},
  pages={e595--e600},
  year={2024},
  publisher={Elsevier}
}

@article{liu2026benchmarking,
  title={Benchmarking large language model-based agent systems for clinical decision tasks},
  author={Liu, Yunsong and Carrero, Zunamys I and Jiang, Xiaofeng and Ferber, Dyke and W{\"o}lflein, Georg and Zhang, Li and Jayabalan, Sanddhya and Lenz, Tim and Hui, Zhouguang and Kather, Jakob Nikolas},
  journal={npj Digital Medicine},
  year={2026},
  publisher={Nature Publishing Group UK London}
}

@article{cifci2023ai,
  title={AI in computational pathology of cancer: improving diagnostic workflows and clinical outcomes?},
  author={Cifci, Didem and Veldhuizen, Gregory P and Foersch, Sebastian and Kather, Jakob Nikolas},
  journal={Annual Review of Cancer Biology},
  volume={7},
  number={1},
  pages={57--71},
  year={2023},
  publisher={Annual Reviews}
}

@article{verghese2023computational,
  title={Computational pathology in cancer diagnosis, prognosis, and prediction--present day and prospects},
  author={Verghese, Gregory and Lennerz, Jochen K and Ruta, Danny and Ng, Wen and Thavaraj, Selvam and Siziopikou, Kalliopi P and Naidoo, Threnesan and Rane, Swapnil and Salgado, Roberto and Pinder, Sarah E and others},
  journal={The Journal of pathology},
  volume={260},
  number={5},
  pages={551--563},
  year={2023},
  publisher={Wiley Online Library}
}

@article{shafique2024preliminary,
  title={A preliminary investigation into search and matching for tumor discrimination in World Health Organization breast taxonomy using deep networks},
  author={Shafique, Abubakr and Gonzalez, Ricardo and Pantanowitz, Liron and Tan, Puay Hoon and Machado, Alberto and Cree, Ian A and Tizhoosh, Hamid R},
  journal={Modern Pathology},
  volume={37},
  number={2},
  pages={100381},
  year={2024},
  publisher={Elsevier}
}

@article{jiang2024transformer,
  title={A transformer-based weakly supervised computational pathology method for clinical-grade diagnosis and molecular marker discovery of gliomas},
  author={Jiang, Rui and Yin, Xiaoxu and Yang, Pengshuai and Cheng, Lingchao and Hu, Juan and Yang, Jiao and Wang, Ying and Fu, Xiaodan and Shang, Li and Li, Liling and others},
  journal={Nature Machine Intelligence},
  volume={6},
  number={8},
  pages={876--891},
  year={2024},
  publisher={Nature Publishing Group UK London}
}

@article{fountzilas2025convergence,
  title={Convergence of evolving artificial intelligence and machine learning techniques in precision oncology},
  author={Fountzilas, Elena and Pearce, Tillman and Baysal, Mehmet A and Chakraborty, Abhijit and Tsimberidou, Apostolia M},
  journal={NPJ Digital Medicine},
  volume={8},
  number={1},
  pages={75},
  year={2025},
  publisher={Nature Publishing Group UK London}
}

@article{yin2024survey,
  title={A survey on multimodal large language models},
  author={Yin, Shukang and Fu, Chaoyou and Zhao, Sirui and Li, Ke and Sun, Xing and Xu, Tong and Chen, Enhong},
  journal={National Science Review},
  volume={11},
  number={12},
  pages={nwae403},
  year={2024},
  publisher={Oxford University Press}
}

@article{caffagni2024revolution,
  title={The revolution of multimodal large language models: A survey},
  author={Caffagni, Davide and Cocchi, Federico and Barsellotti, Luca and Moratelli, Nicholas and Sarto, Sara and Baraldi, Lorenzo and Cornia, Marcella and Cucchiara, Rita},
  journal={Findings of the association for computational linguistics: ACL 2024},
  pages={13590--13618},
  year={2024}
}

@article{liu2023visual,
  title={Visual instruction tuning},
  author={Liu, Haotian and Li, Chunyuan and Wu, Qingyang and Lee, Yong Jae},
  journal={Advances in neural information processing systems},
  volume={36},
  pages={34892--34916},
  year={2023}
}

@inproceedings{liu2025unveiling,
  title={Unveiling the ignorance of mllms: Seeing clearly, answering incorrectly},
  author={Liu, Yexin and Liang, Zhengyang and Wang, Yueze and Wu, Xianfeng and Tang, Feilong and He, Muyang and Li, Jian and Liu, Zheng and Yang, Harry and Lim, Sernam and others},
  booktitle={Proceedings of the Computer Vision and Pattern Recognition Conference},
  pages={9087--9097},
  year={2025}
}

@article{hou2024memory,
  title={Memory-augmented multimodal llms for surgical vqa via self-contained inquiry},
  author={Hou, Wenjun and Cheng, Yi and Xu, Kaishuai and Hu, Yan and Li, Wenjie and Liu, Jiang},
  journal={arXiv preprint arXiv:2411.10937},
  year={2024}
}

@article{bei2026mem,
  title={Mem-gallery: Benchmarking multimodal long-term conversational memory for mllm agents},
  author={Bei, Yuanchen and Wei, Tianxin and Ning, Xuying and Zhao, Yanjun and Liu, Zhining and Lin, Xiao and Zhu, Yada and Hamann, Hendrik and He, Jingrui and Tong, Hanghang},
  journal={arXiv preprint arXiv:2601.03515},
  year={2026}
}

@incollection{baddeley2025working,
  title={Working memory},
  author={Baddeley, Alan},
  booktitle={Memory},
  pages={101--152},
  year={2025},
  publisher={Routledge}
}

@inproceedings{hamza2025llava,
  title={Llava needs more knowledge: Retrieval augmented natural language generation with knowledge graph for explaining thoracic pathologies},
  author={Hamza, Ameer and Ahn, Yong Hyun and Lee, Sungyoung and Kim, Seong Tae and others},
  booktitle={Proceedings of the AAAI Conference on Artificial Intelligence},
  volume={39},
  number={3},
  pages={3311--3319},
  year={2025}
}

@article{jabal2025open,
  title={Open-weight language models and retrieval-augmented generation for automated structured data extraction from diagnostic reports: assessment of approaches and parameters},
  author={Jabal, Mohamed Sobhi and Warman, Pranav and Zhang, Jikai and Gupta, Kartikeye and Jain, Ayush and Mazurowski, Maciej and Wiggins, Walter and Magudia, Kirti and Calabrese, Evan},
  journal={Radiology: Artificial Intelligence},
  volume={7},
  number={3},
  pages={e240551},
  year={2025},
  publisher={Radiological Society of North America}
}

@article{an2025cognitive,
  title={Cognitive Workspace: Active Memory Management for LLMs--An Empirical Study of Functional Infinite Context},
  author={An, Tao},
  journal={arXiv preprint arXiv:2508.13171},
  year={2025}
}

@article{huang2026ama,
  title={AMA: Adaptive Memory via Multi-Agent Collaboration},
  author={Huang, Weiquan and Wang, Zixuan and Lin, Hehai and Wang, Sudong and Xu, Bo and Li, Qian and Zhu, Beier and Yang, Linyi and Qin, Chengwei},
  journal={arXiv preprint arXiv:2601.20352},
  year={2026}
}

@article{jiang2026synapse,
  title={SYNAPSE: Empowering LLM Agents with Episodic-Semantic Memory via Spreading Activation},
  author={Jiang, Hanqi and Chen, Junhao and Pan, Yi and Chen, Ling and You, Weihang and Zhou, Yifan and Zhang, Ruidong and Sikora, Andrea and Zhao, Lin and Abate, Yohannes and others},
  journal={arXiv preprint arXiv:2601.02744},
  year={2026}
}

@inproceedings{hou2025synapticrag,
  title={SynapticRAG: Enhancing Temporal Memory Retrieval in Large Language Models through Synaptic Mechanisms},
  author={Hou, Yuki and Tamoto, Haruki and Zhao, Qinghua and Miyashita, Homei},
  booktitle={Findings of the Association for Computational Linguistics: ACL 2025},
  pages={20422--20436},
  year={2025}
}

@inproceedings{yoshida2025if,
  title={If Attention Serves as a Cognitive Model of Human Memory Retrieval, What is the Plausible Memory Representation?},
  author={Yoshida, Ryo and Isono, Shinnosuke and Kajikawa, Kohei and Someya, Taiga and Sugimoto, Yushi and Oseki, Yohei},
  booktitle={Proceedings of the 63rd Annual Meeting of the Association for Computational Linguistics (Volume 1: Long Papers)},
  pages={9795--9812},
  year={2025}
}

@article{dong2025towards,
  title={Towards large language models with human-like episodic memory},
  author={Dong, Cody V and Lu, Qihong and Norman, Kenneth A and Michelmann, Sebastian},
  journal={Trends in Cognitive Sciences},
  year={2025},
  publisher={Elsevier}
}

@inproceedings{sun2024pathmmu,
  title={Pathmmu: A massive multimodal expert-level benchmark for understanding and reasoning in pathology},
  author={Sun, Yuxuan and Wu, Hao and Zhu, Chenglu and Zheng, Sunyi and Chen, Qizi and Zhang, Kai and Zhang, Yunlong and Wan, Dan and Lan, Xiaoxiao and Zheng, Mengyue and others},
  booktitle={European Conference on Computer Vision},
  pages={56--73},
  year={2024},
  organization={Springer}
}

@article{chen2025pathagent,
  title={Pathagent: Toward interpretable analysis of whole-slide pathology images via large language model-based agentic reasoning},
  author={Chen, Jingyun and Cai, Linghan and Wang, Zhikang and Huang, Yi and Jiang, Songhan and Huang, Shenjin and Wang, Hongpeng and Zhang, Yongbing},
  journal={arXiv preprint arXiv:2511.17052},
  year={2025}
}

@article{liu2025pathmind,
  title={PathMind: A Retrieve-Prioritize-Reason Framework for Knowledge Graph Reasoning with Large Language Models},
  author={Liu, Yu and Lin, Xixun and Shang, Yanmin and Li, Yangxi and Wang, Shi and Cao, Yanan},
  journal={arXiv preprint arXiv:2511.14256},
  year={2025}
}

@article{lu2024multimodal,
  title={A multimodal generative AI copilot for human pathology},
  author={Lu, Ming Y and Chen, Bowen and Williamson, Drew FK and Chen, Richard J and Zhao, Melissa and Chow, Aaron K and Ikemura, Kenji and Kim, Ahrong and Pouli, Dimitra and Patel, Ankush and others},
  journal={Nature},
  volume={634},
  number={8033},
  pages={466--473},
  year={2024},
  publisher={Nature Publishing Group UK London}
}

@article{chen2025evidence,
  title={Evidence-based diagnostic reasoning with multi-agent copilot for human pathology},
  author={Chen, Chengkuan and Weishaupt, Luca L and Williamson, Drew FK and Chen, Richard J and Ding, Tong and Chen, Bowen and Vaidya, Anurag and Le, Long Phi and Jaume, Guillaume and Lu, Ming Y and others},
  journal={arXiv preprint arXiv:2506.20964},
  year={2025}
}

@article{zhou2025mllm4pue,
  title={Mllm4pue: toward universal embeddings in computational pathology through multimodal llms},
  author={Zhou, Qifeng and Dang, Thao M and Zhong, Wenliang and Guo, Yuzhi and Ma, Hehuan and Na, Saiyang and Huang, Junzhou},
  journal={arXiv e-prints},
  pages={arXiv--2502},
  year={2025}
}

@article{chen2024metapath,
  title={MetaPath Chat: multimodal generative artificial intelligence chatbot for clinical pathology},
  author={Chen, Haizhu and Lin, Ruichong and Yunfang, Yu},
  journal={MedComm},
  volume={5},
  number={10},
  pages={e769},
  year={2024}
}

@article{allaberdiev2026chestxgen,
  title={Chestxgen: Dynamic Memory-Augmented Vision-Language Transformer with Context-Aware Gating for Radiology Report Generation},
  author={Allaberdiev, Sharofiddin and Khan, Asif and Mamarasulov, Sardor and Chen, Xiaojun},
  journal={Journal of Artificial Intelligence and Soft Computing Research},
  volume={16},
  number={1},
  pages={55--72},
  year={2026},
  publisher={De Gruyter Poland}
}

@article{cabalo2024differential,
  title={Differential reorganization of episodic and semantic memory systems in epilepsy-related mesiotemporal pathology},
  author={Cabalo, Donna Gift and DeKraker, Jordan and Royer, Jessica and Xie, Ke and Tavakol, Shahin and Rodr{\'\i}guez-Cruces, Ra{\'u}l and Bernasconi, Andrea and Bernasconi, Neda and Weil, Alexander and Pana, Raluca and others},
  journal={Brain},
  volume={147},
  number={11},
  pages={3918--3932},
  year={2024},
  publisher={Oxford University Press UK}
}

@article{sheng2025top,
  title={Top-down attention and Alzheimer’s pathology affect cortical selectivity during learning, influencing episodic memory in older adults},
  author={Sheng, Jintao and Trelle, Alexandra N and Romero, America and Park, Jennifer and Tran, Tammy T and Sha, Sharon J and Andreasson, Katrin I and Wilson, Edward N and Mormino, Elizabeth C and Wagner, Anthony D},
  journal={Science advances},
  volume={11},
  number={24},
  pages={eads4206},
  year={2025},
  publisher={American Association for the Advancement of Science}
}

@article{yang2022long,
  title={Long-term exercise pre-training attenuates Alzheimer’s disease--related pathology in a transgenic rat model of Alzheimer’s disease},
  author={Yang, Luodan and Wu, Chongyun and Li, Yong and Dong, Yan and Wu, Celeste Yin-Chieh and Lee, Reggie Hui-Chao and Brann, Darrell W and Lin, Hung Wen and Zhang, Quanguang},
  journal={Geroscience},
  volume={44},
  number={3},
  pages={1457--1477},
  year={2022},
  publisher={Springer}
}

@article{minhas2024restoring,
  title={Restoring hippocampal glucose metabolism rescues cognition across Alzheimer’s disease pathologies},
  author={Minhas, Paras S and Jones, Jeffrey R and Latif-Hernandez, Amira and Sugiura, Yuki and Durairaj, Aarooran S and Wang, Qian and Mhatre, Siddhita D and Uenaka, Takeshi and Crapser, Joshua and Conley, Travis and others},
  journal={Science},
  volume={385},
  number={6711},
  pages={eabm6131},
  year={2024},
  publisher={American Association for the Advancement of Science}
}

@article{blackmore2023ultrasound,
  title={Ultrasound as a versatile tool for short-and long-term improvement and monitoring of brain function},
  author={Blackmore, Daniel G and Razansky, Daniel and G{\"o}tz, J{\"u}rgen},
  journal={Neuron},
  volume={111},
  number={8},
  pages={1174--1190},
  year={2023},
  publisher={Elsevier}
}

@article{akhtar2024types,
  title={Types of memory, dementia, Alzheimer’s disease, and their various pathological cascades as targets for potential pharmacological drugs},
  author={Akhtar, Ansab and Singh, Siddharth and Kaushik, Ravinder and Awasthi, Rajendra and Behl, Tapan},
  journal={Ageing Research Reviews},
  volume={96},
  pages={102289},
  year={2024},
  publisher={Elsevier}
}

@article{karakatsani2023focused,
  title={Focused ultrasound mitigates pathology and improves spatial memory in Alzheimer's mice and patients},
  author={Karakatsani, Maria Eleni and Ji, Robin and Murillo, Maria F and Kugelman, Tara and Kwon, Nancy and Lao, Yeh-Hsing and Liu, Keyu and Pouliopoulos, Antonios N and Honig, Lawrence S and Duff, Karen E and others},
  journal={Theranostics},
  volume={13},
  number={12},
  pages={4102},
  year={2023}
}

@article{manchanda2023intravenous,
  title={Intravenous treatment with a molecular chaperone designed against $\beta$-amyloid toxicity improves Alzheimer’s disease pathology in mouse models},
  author={Manchanda, Shaffi and Galan-Acosta, Lorena and Abelein, Axel and Tambaro, Simone and Chen, Gefei and Nilsson, Per and Johansson, Jan},
  journal={Molecular Therapy},
  volume={31},
  number={2},
  pages={487--502},
  year={2023},
  publisher={Elsevier}
}

@article{yu2025ypathrag,
  title={YpathRAG: A Retrieval-Augmented Generation Framework and Benchmark for Pathology},
  author={Yu, Deshui and Wang, Yizhi and Jin, Saihui and Zhu, Taojie and Zeng, Fanyi and Qian, Wen and Huang, Zirui and Ouyang, Jingli and Li, Jiameng and Song, Zhen and others},
  journal={arXiv preprint arXiv:2510.08603},
  year={2025}
}

@article{fink2025retrieval,
  title={Retrieval-augmented generation with large language models in radiology: from theory to practice},
  author={Fink, Anna and Rau, Alexander and Reisert, Marco and Bamberg, Fabian and Russe, Maximilian F},
  journal={Radiology: Artificial Intelligence},
  volume={7},
  number={4},
  pages={e240790},
  year={2025},
  publisher={Radiological Society of North America}
}

@article{redline2023placental,
  title={Placental pathology is necessary to understand common pregnancy complications and achieve an improved taxonomy of obstetrical disease},
  author={Redline, Raymond W and Roberts, Drucilla J and Parast, Mana M and Ernst, Linda M and Morgan, Terry K and Greene, Michael F and Gyamfi-Bannerman, Cynthia and Louis, Judette M and Maltepe, Emin and Mestan, Karen K and others},
  journal={American journal of obstetrics and gynecology},
  volume={228},
  number={2},
  pages={187--202},
  year={2023},
  publisher={Elsevier}
}

@article{romero2022toward,
  title={Toward a new taxonomy of obstetrical disease: improved performance of maternal blood biomarkers for the great obstetrical syndromes when classified according to placental pathology},
  author={Romero, Roberto and Jung, Eunjung and Chaiworapongsa, Tinnakorn and Erez, Offer and Gudicha, Dereje W and Kim, Yeon Mee and Kim, Jung-Sun and Kim, Bomi and Kusanovic, Juan Pedro and Gotsch, Francesca and others},
  journal={American journal of obstetrics and gynecology},
  volume={227},
  number={4},
  pages={615--e1},
  year={2022},
  publisher={Elsevier}
}

@article{gabitto2024integrated,
  title={Integrated multimodal cell atlas of Alzheimer’s disease},
  author={Gabitto, Mariano I and Travaglini, Kyle J and Rachleff, Victoria M and Kaplan, Eitan S and Long, Brian and Ariza, Jeanelle and Ding, Yi and Mahoney, Joseph T and Dee, Nick and Goldy, Jeff and others},
  journal={Nature neuroscience},
  volume={27},
  number={12},
  pages={2366--2383},
  year={2024},
  publisher={Nature Publishing Group US New York}
}

@article{gosai2025beyond,
  title={Beyond Diagnosis: Evaluating Multimodal LLMs for Pathology Localization in Chest Radiographs},
  author={Gosai, Advait and Kavishwar, Arun and McNamara, Stephanie L and Samineni, Soujanya and Umeton, Renato and Chowdhury, Alexander and Lotter, William},
  journal={arXiv preprint arXiv:2509.18015},
  year={2025}
}

@article{li2025multi,
  title={Multi-modal foundation models for computational pathology: a survey},
  author={Li, Dong and Wan, Guihong and Wu, Xintao and Wu, Xinyu and Chen, Xiaohui and He, Yi and Chen, Zhong and Sorger, Peter K and Zhao, Chen},
  journal={Transactions on machine learning research},
  volume={2025},
  pages={5715},
  year={2025}
}

@article{ullah2025multimodal,
  title={Multimodal generative AI for anatomic pathology—A review of current applications to envisage the future direction},
  author={Ullah, Ehsan and Baig, Mirza Mansoor and Waqas, Asim and Rasool, Ghulam and Singh, Rajendra and Shandilya, Ashwinikumar and GholamHossieni, Hamid and Parwani, Anil V},
  journal={Advances in anatomic pathology},
  pages={10--1097},
  year={2025},
  publisher={LWW}
}

@misc{bian2026biomedical,
  title={Biomedical multimodal large language models: From model-centric development to clinically grounded evaluation and integration},
  author={Bian, Jiang and Peng, Yifan and Mendonca, Eneida and Banerjee, Imon and Xu, Hua},
  journal={Journal of Biomedical Informatics},
  pages={105023},
  year={2026},
  publisher={Elsevier}
}
\bibliographystyle{plain}

\appendix
\section{Long-term Memory Construction}
\label{KG Construction}

To construct the pathology-oriented knowledge graph, we design an evidence-driven
deep-search pipeline that automatically mines disease-relevant knowledge from
PubMed abstracts (As shown in Algorithm~\ref{algo:pathology_kg_construction}). Given a predefined disease query, the pipeline first retrieves
PubMed identifiers and fetches their corresponding XML records through the
NCBI Entrez interface. Abstracts without valid textual content are discarded,
and duplicate abstracts are removed by normalized hashing. Each remaining
abstract is processed by a large language model using a constrained pathology
schema, including disease entities, anatomical sites, histology, morphologic
features, immunohistochemical markers, molecular alterations, serum markers,
and diagnostic clues. The extraction module is instructed to output only
explicitly supported information with evidence spans and confidence scores.
To improve reliability, each extracted relation is further normalized,
confidence-calibrated, and filtered by a minimum confidence threshold before
being inserted into the knowledge graph. The final graph is saved both as
plain triples and as rich evidence-aware edges, while disease-centric node
memories and feature-to-disease indices are exported for subsequent retrieval
and model guidance.

\begin{algorithm}[H]
\DontPrintSemicolon
\SetAlgoLined
\KwIn{Disease-specific PubMed query $Q$;
      maximum number of articles $N$;
      confidence threshold $\tau$;
      LLM extraction schema $\mathcal{S}$}
\KwOut{Pathology knowledge graph $\mathcal{G}=(\mathcal{V},\mathcal{E})$;
       triples $\mathcal{T}$;
       disease memory $\mathcal{M}_{d}$;
       feature index $\mathcal{M}_{f}$}

Initialize empty pathology KG $\mathcal{G}$ and triple set $\mathcal{T}$\;
Initialize seen abstract hash set $\mathcal{H}$\;

\tcc{PubMed deep search}
$\mathcal{P} \leftarrow \mathrm{PubMedSearch}(Q,N)$\;
$\mathcal{A} \leftarrow \mathrm{PubMedFetch}(\mathcal{P})$\;

\ForEach{article $a \in \mathcal{A}$}{
  Extract PMID $m$ and abstract text $x$ from $a$\;
  \If{$x$ is empty}{
    \textbf{continue}\;
  }

  \tcc{Abstract-level deduplication}
  $h \leftarrow \mathrm{Hash}(\mathrm{Normalize}(x))$\;
  \If{$h \in \mathcal{H}$}{
    \textbf{continue}\;
  }
  $\mathcal{H} \leftarrow \mathcal{H} \cup \{h\}$\;

  \tcc{Schema-constrained LLM extraction}
  $r \leftarrow \mathrm{LLMExtract}(x,\mathcal{S})$\;
  Save raw extraction $(m,r)$ for audit and retry\;

  \If{$r$ is invalid}{
    \textbf{continue}\;
  }

  Let $d$ be the normalized disease name in $r$\;
  \If{$d$ is empty}{
    \textbf{continue}\;
  }

  \tcc{Relation normalization and graph insertion}
  \ForEach{candidate item $(s,p,o,e,c)$ extracted from $r$}{
    $p' \leftarrow \mathrm{SlugRelation}(p)$\;
    $c' \leftarrow \mathrm{CalibrateConfidence}(c,e)$\;
    $e' \leftarrow \mathrm{ClipEvidence}(e)$\;

    \If{$c' \geq \tau$ and $s,o,p'$ are valid}{
      Add triple $(s,p',o)$ into $\mathcal{T}$\;
      Add rich edge $(s,p',o,m,c',e')$ into $\mathcal{E}$\;
      Update node set $\mathcal{V} \leftarrow \mathcal{V} \cup \{s,o\}$\;
    }
  }
}

\tcc{Memory export}
Save $\mathcal{T}$ as \texttt{triples.tsv}\;
Save evidence-aware edges $\mathcal{E}$ as \texttt{edges.jsonl}\;
$\mathcal{M}_{d} \leftarrow \mathrm{ExportDiseaseNodes}(\mathcal{T})$\;
$\mathcal{M}_{f} \leftarrow \mathrm{ExportFeatureIndex}(\mathcal{T})$\;

\Return $\mathcal{G}, \mathcal{T}, \mathcal{M}_{d}, \mathcal{M}_{f}$\;

\caption{Pseudocode of the pathology knowledge graph construction pipeline.}
\label{algo:pathology_kg_construction}
\end{algorithm}

\section{More Implementation Details}
\label{appendix More Implementation Details}

\subsection{Dataset \& Processing}
\label{appendix Dataset Processing}
Experiments are mainly conducted on \textbf{WSI-Bench}, with additional zero-shot studies on independent datasets to evaluate generalization ability.
\textbf{In-domain benchmark.}
We use \textbf{WSI-Bench} as the core benchmark for both model development and quantitative evaluation. Derived from TCGA, it focuses on pathology-specific visual reasoning over whole-slide images (WSIs). The benchmark consists of 9,850 WSIs spanning 30 tumor types and includes 179,569 associated VQA instances.
The annotations are designed to evaluate four key aspects of pathological intelligence:
(1) morphology understanding,
(2) diagnostic assessment,
(3) therapy-related reasoning, and
(4) report-level text generation.
Both generative and multiple-choice answer formats are included.
We strictly follow the official data split without modification. Specifically, 9,642 WSIs with 175,450 QA pairs are used for training, while 208 WSIs with 4,119 QA pairs are reserved for evaluation. All reported metrics follow the original benchmark protocol, including WSI-P, WSI-R, classification accuracy, BLEU, and ROUGE.
\textbf{Out-of-domain benchmarks.}
To evaluate generalization to unseen data distributions, we conduct zero-shot experiments on three independent datasets: \textit{WSI-VQA}~\cite{chen2024wsi}, \textit{SlideBench-VQA (BCNB)}~\cite{chen2025slidechat}, and \textit{CPTAC-NSCLC}.
Following prior work, we apply data-cleaning procedures to remove samples that are not clinically well-posed or cannot be answered based on WSI content. This includes questions involving prognosis prediction, unavailable biomarker information, or three-dimensional lesion size estimation.
Importantly, these external datasets are used strictly for evaluation purposes, and no additional fine-tuning or adaptation is performed.

\subsection{Implementation Details}
\label{appendix Implementation Details}
\textbf{WSI Representation Pipeline.}
Given the gigapixel nature of whole-slide images, we adopt a hierarchical processing strategy to balance efficiency and contextual awareness. Each WSI is first divided into non-overlapping tiles of size 256$\times$256 at a predefined magnification level.
These tiles are independently encoded using a visual backbone to produce patch-level feature embeddings, following WSI-LLaVA~\cite{liang2025wsi}.
To capture global structural information, the patch embeddings are aggregated and passed through a LongNet-based transformer. This module is designed to model long-range dependencies across the entire slide, enabling effective representation of spatial relationships that extend beyond local regions. The final visual representation is projected into the language embedding space via a trainable linear layer, which serves as the interface between vision and language modalities.
\textbf{Language Modeling and Multimodal Fusion.}
We leverage a pretrained large language model (LLM) as the textual backbone for downstream reasoning and generation tasks. The projected visual features are aligned with token embeddings and injected into the LLM through a multimodal projector. This design allows the model to condition text generation on slide-level visual context. Depending on the training stage, different subsets of parameters are frozen to stabilize optimization and reduce computational cost.
\textbf{Multi-Stage Optimization.}
Training is organized into three sequential stages:
\ding{182} \textbf{Cross-Modal Pre-alignment.} We first perform contrastive learning between slide-level visual embeddings and corresponding pathology reports to establish a shared embedding space.
\ding{183} \textbf{Projection Adaptation.} Next, we connect the pretrained visual encoder with the LLM through a projection layer. During this stage, both the visual encoder and LLM remain frozen, and only the projection parameters are updated.
\ding{184} \textbf{Instruction Fine-tuning.} Finally, we conduct task-specific training on WSI-Bench. The projection layer and LLM are jointly optimized, while the visual encoder is kept fixed to preserve learned visual representations.
\emph{\textbf{Evaluation Metrics.}}
We evaluate model performance from both linguistic and clinical perspectives. For open-ended generation tasks, BLEU, ROUGE-L, and METEOR are used to assess lexical overlap and semantic similarity with reference reports. To better reflect pathology-specific correctness, we additionally employ WSI-P and WSI-R, which measure the accuracy of predicted findings and their consistency with visual evidence, respectively.
\textbf{Training Configuration.}
All experiments are conducted using DeepSpeed on 4 NVIDIA A100 GPUs. Optimization is performed with AdamW, using an initial learning rate of $2\times10^{-4}$ and a cosine decay schedule with a warmup ratio of 0.03. The model is trained for 3 epochs with a per-device batch size of 32.
For parameter-efficient adaptation, we adopt LoRA with rank $r=128$ and scaling factor $\alpha=256$. The multimodal projection layer is trained with a smaller learning rate of $2\times10^{-5}$ to ensure stable alignment between modalities.
\section{Evaluation Prompts}
\label{appendix Evaluation Prompts}

To enable fine-grained and more reliable evaluation of model-generated pathology outputs, we design a multi-stage prompting pipeline for both \textbf{WSI-R (WSI-R)} and \textbf{WSI-P (WSI-P)}. Our evaluation operates at the \textit{claim level}, where free-form text is decomposed into atomic factual units and subsequently assessed using large language models (LLMs).
\emph{\ding{182} \textbf{Claim Extraction.}}
We first transform long-form pathology text into structured atomic claims.
\textbf{WSI-R.}
For WSI-R, claims are extracted from model-generated diagnostic text. The prompt instructs the LLM to decompose the text into granular, non-overlapping claims while preserving semantic coherence. Closely related information is kept within the same claim to avoid breaking contextual dependencies. The model is required to avoid omission or duplication and to output results strictly as a list of claims.
\textbf{WSI-P.}
For WSI-P, claim extraction is conditioned on question-answer pairs. The prompt additionally filters out information that is not directly relevant to the question and retains only claims that explicitly address the query. This ensures that subsequent evaluation focuses on answer-specific correctness rather than general content.
\emph{\ding{183} \textbf{Claim-level Evaluation.}}
After claim extraction, we evaluate each claim using task-specific prompts.
\textbf{WSI-R Evaluation.}
In WSI-R, each model-generated claim is evaluated against the ground truth answer. The LLM acts as an impartial judge and assigns a graded relevance score along with a natural language explanation.
The scoring rubric is defined as follows:
\textbf{1.0}: Fully supported by the ground truth
\textbf{0.7}: Mostly supported with minor omissions
\textbf{0.3}: Partially supported with significant gaps
\textbf{0.0}: Not supported
Claims that are not addressed in the ground truth are ignored to avoid penalizing missing coverage.
\textbf{WSI-P Evaluation.}
In WSI-P, evaluation focuses on factual correctness. Each ground truth claim is compared against the model response, and the LLM assigns a correctness score with explanation.
The scoring rubric is:
\textbf{1.0}: Completely correct
\textbf{0.7}: Mostly correct
\textbf{0.3}: Contains core factual errors
\textbf{0.0}: Completely incorrect
This formulation emphasizes clinically meaningful correctness, including subtle diagnostic inaccuracies.
\emph{\ding{184} \textbf{Output Format.}}
Both evaluation prompts require the LLM to output results as a list of structured entries, where each entry contains:
the claim,
an explanation,
and a numerical score.
This standardized format facilitates reliable parsing and downstream aggregation.
\emph{\ding{185} \textbf{Score Aggregation}}
Given claim-level scores, we compute:
\textbf{Per-sample score}: average across all claims in a sample,
\textbf{Type-level score}: average within each metadata category,
\textbf{Overall score}: average across all samples.
Invalid or missing claims are excluded from aggregation to ensure robustness.
\emph{\ding{186} \textbf{Design Rationale}}
Our prompting design is guided by three principles:
\textbf{Decomposition}: Breaking complex pathology text into atomic claims reduces ambiguity.
\textbf{Separation of Concerns}: WSI-R evaluates relevance (coverage), while WSI-P evaluates correctness.
\textbf{Graded Scoring}: Continuous scores provide more nuanced evaluation than binary metrics.

\section{Compute Reporting}
\label{Compute Reporting}
All experiments were conducted using \texttt{DeepSpeed} on 4 NVIDIA A100 GPUs. The model was optimized with AdamW using an initial learning rate of $2\times10^{-4}$, together with a cosine decay learning rate schedule and a warmup ratio of 0.03. Training was performed for 3 epochs with a per-device batch size of 32.
For parameter-efficient fine-tuning, we adopted LoRA with rank $r=128$ and scaling factor $\alpha=256$. To stabilize multimodal alignment, the multimodal projection layer was optimized with a smaller learning rate of $2\times10^{-5}$.
The overall compute budget was moderate due to the use of parameter-efficient adaptation. Compared with full-parameter fine-tuning, LoRA significantly reduced the number of trainable parameters and GPU memory consumption, enabling all experiments to be completed on a 4-GPU setup. In addition, mixed-precision training and distributed optimization provided by \texttt{DeepSpeed} were used to further improve computational efficiency and reduce training overhead.

\section{More Qualitative Case Study}
\label{appendix More Qualitative Case Study}
\begin{figure}[!h]
  \centering
  \includegraphics[width=1\linewidth]{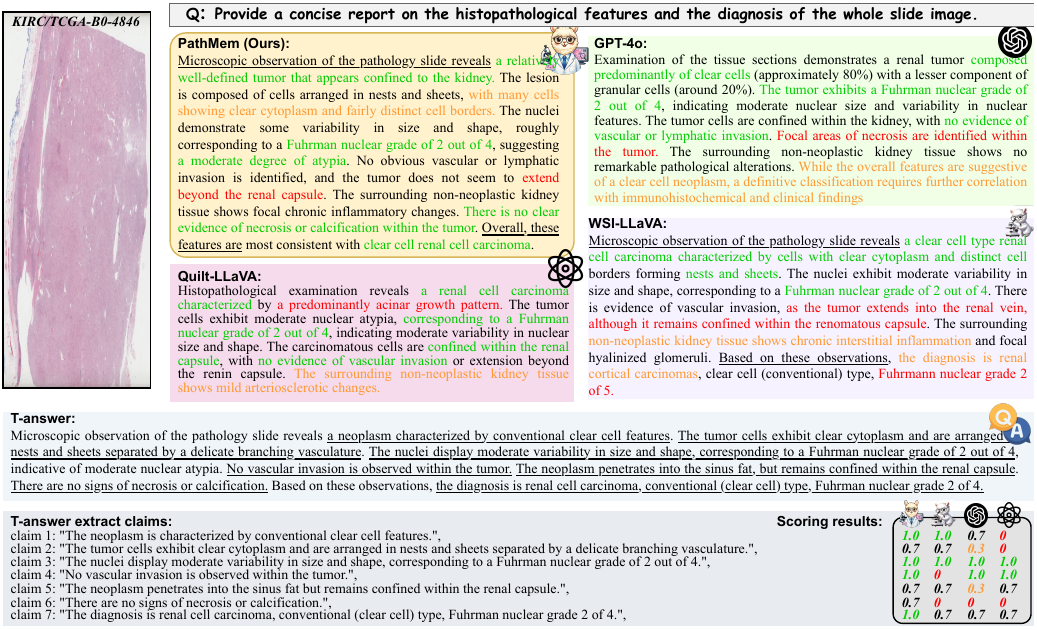}
  \caption{Qualitative comparison of generated reports by our method and three baseline approaches on the report generation task.
(Red highlights denote incorrect content, while green highlights denote correct content, and orange highlights denote missing content from T-answer.)}
  \label{case12}
\end{figure}

\begin{figure}[!h]
\centering
\includegraphics[width=1\textwidth]{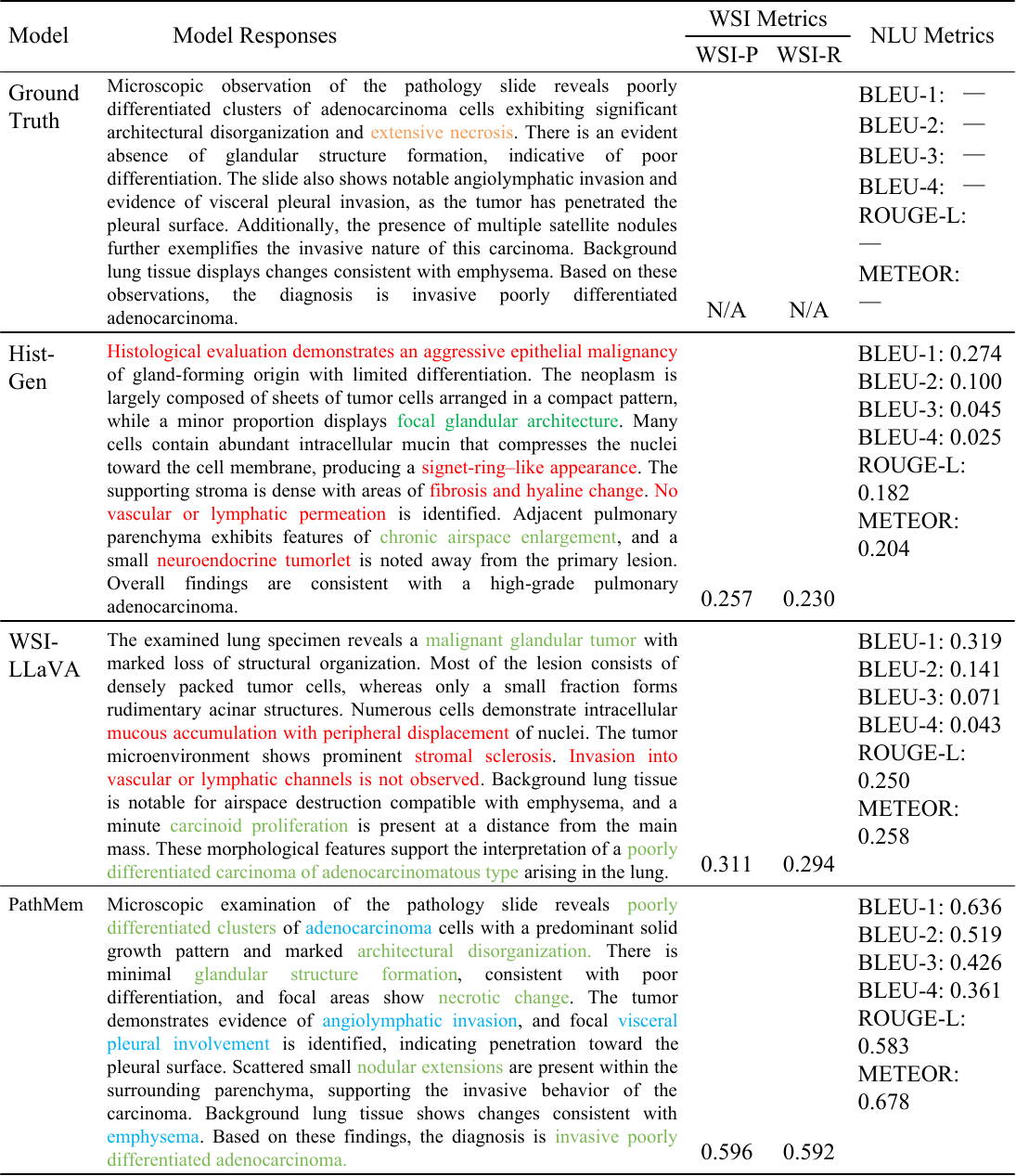}
\caption{Comparison of WSI-based and NLU-based evaluations. Green indicates agreement with the reference, red denotes deviations, orange marks missing ground-truth content, underlined text reflects template-style language, and blue highlights knowledge graph–grounded concepts.}
\label{KG-activationV2}
\end{figure}

\begin{figure}[!h]
  \centering
  \includegraphics[width=1\linewidth]{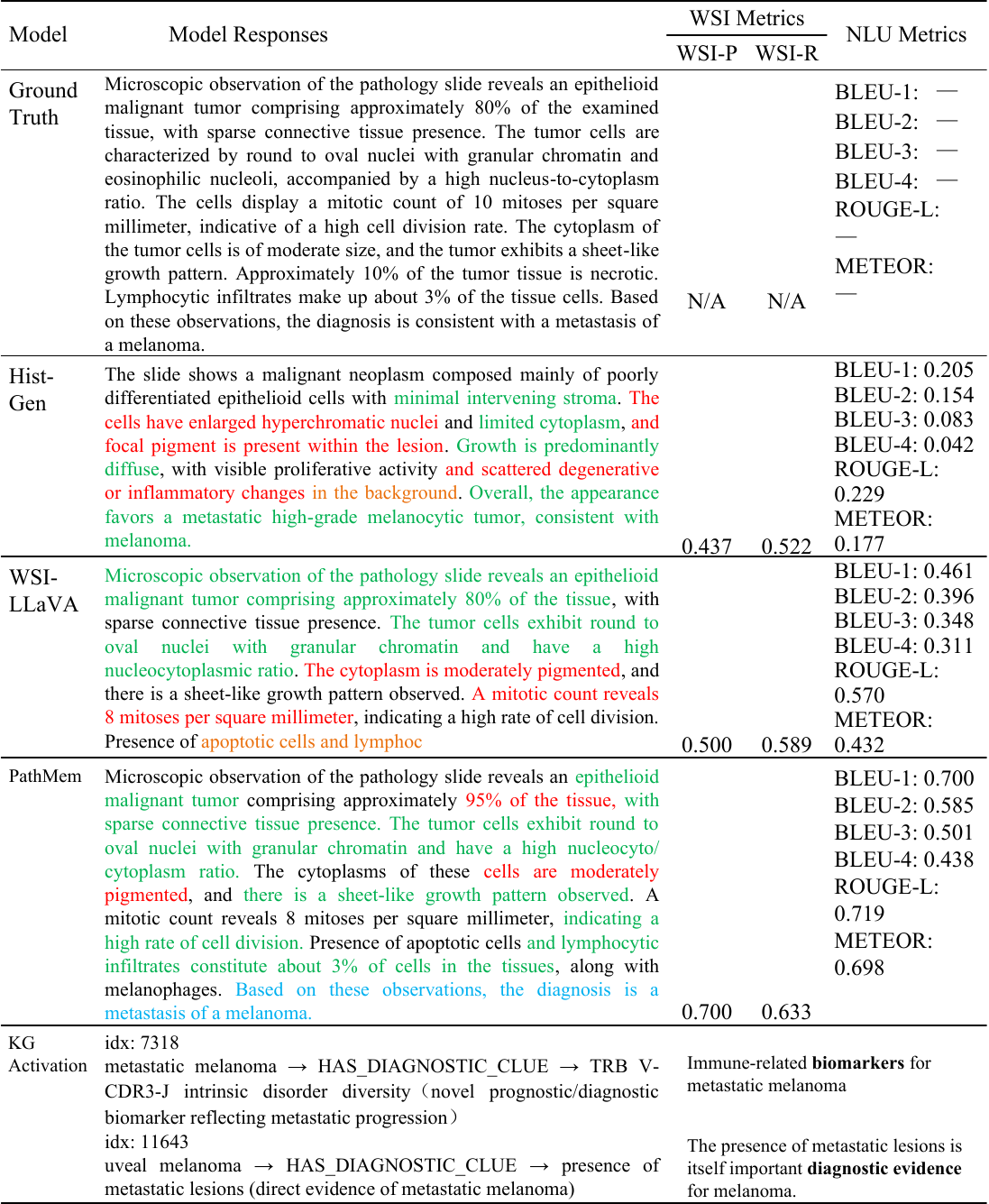}
  \caption{Comparison of WSI-based and NLU-based evaluations. Green indicates agreement with the reference, red denotes deviations, orange marks missing ground-truth content, underlined text reflects template-style language, and blue highlights knowledge graph–grounded concepts.}
  \label{case1}
\end{figure}

\begin{figure}[!h]
  \centering
  \includegraphics[width=1\linewidth]{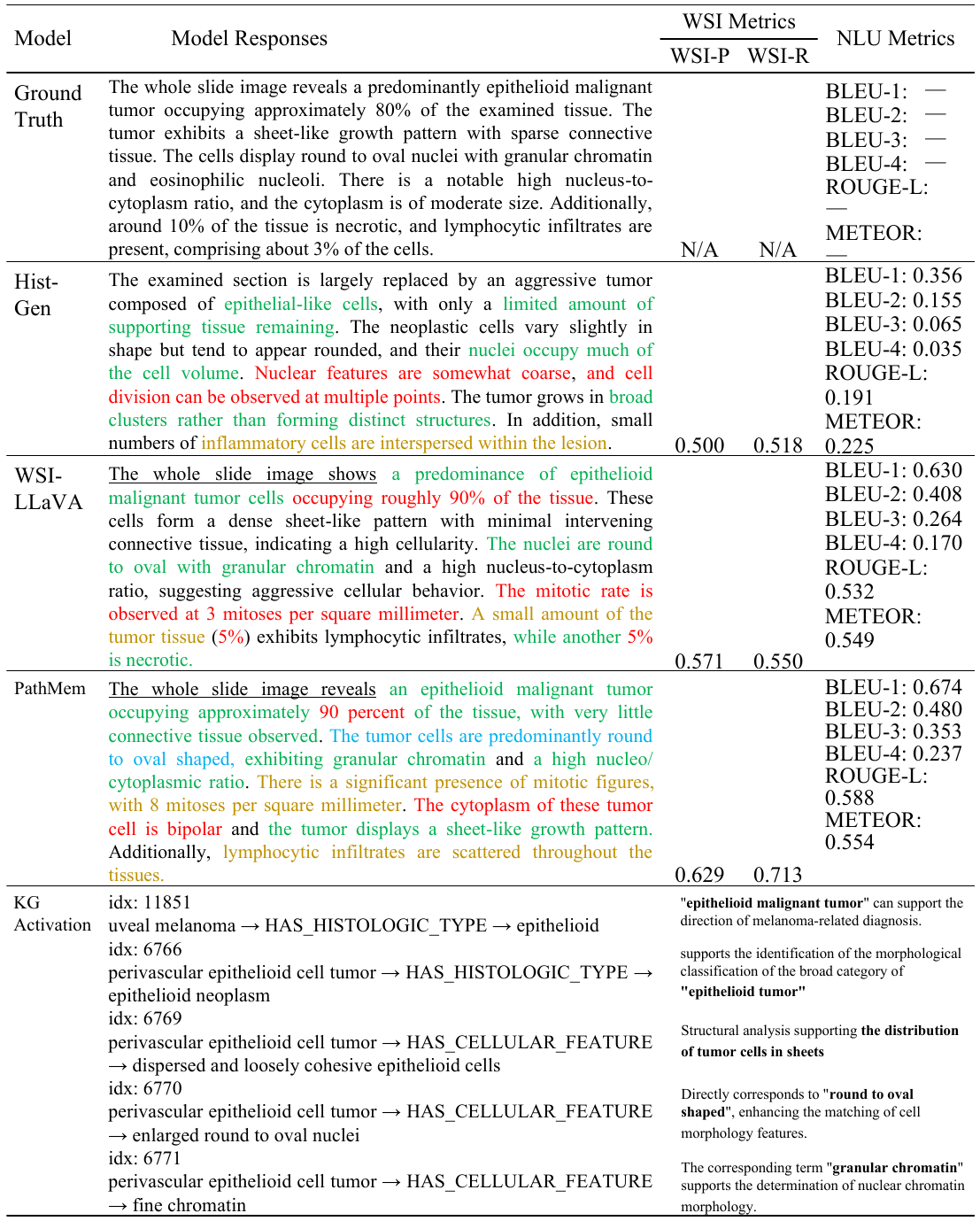}
  \caption{Comparison of WSI-based and NLU-based evaluations. Green indicates agreement with the reference, red denotes deviations, orange marks missing ground-truth content, underlined text reflects template-style language, and blue highlights knowledge graph–grounded concepts.}
  \label{case3}
\end{figure}

\begin{figure}[!h]
  \centering
  \includegraphics[width=1\linewidth]{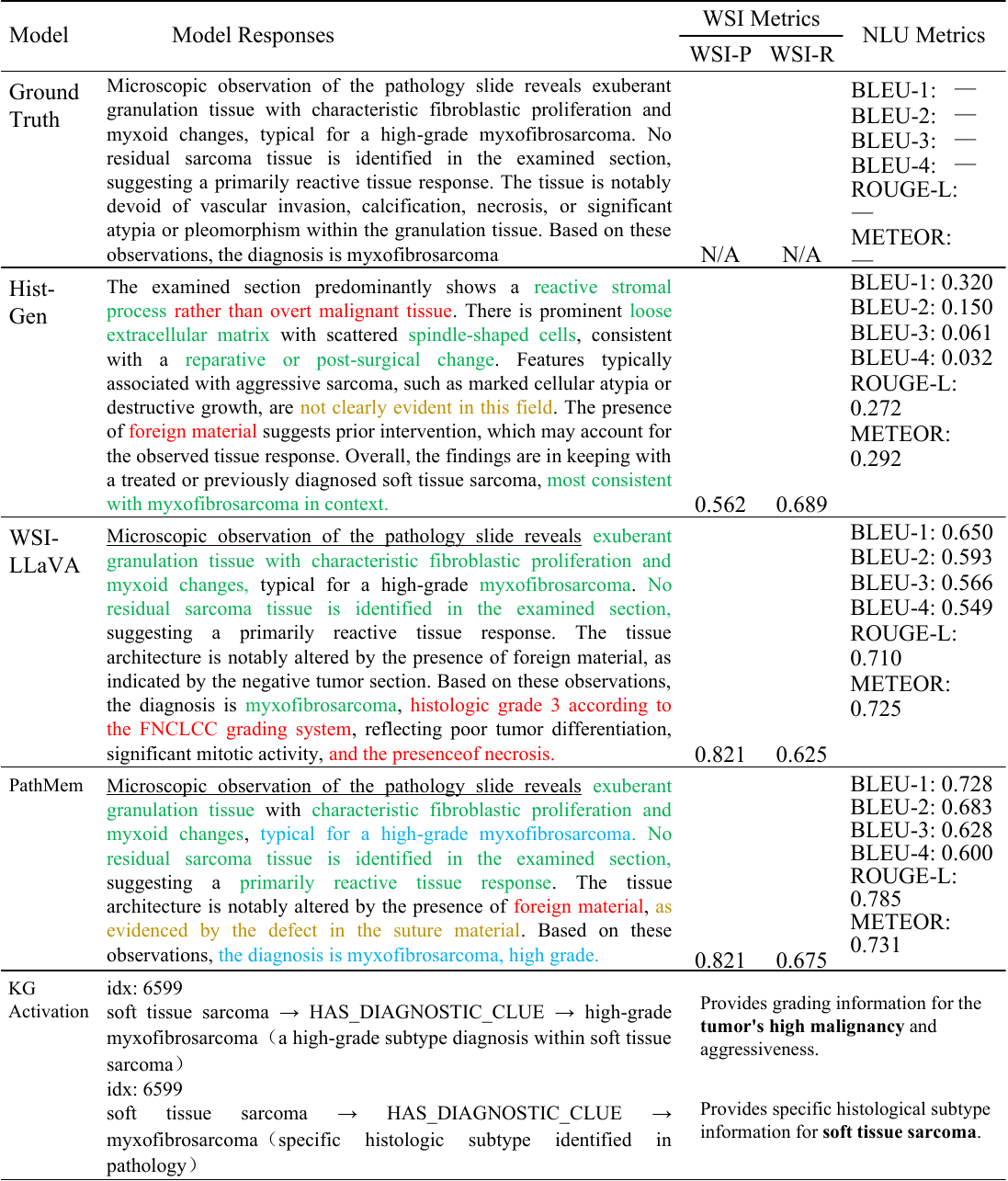}
  \caption{Comparison of WSI-based and NLU-based evaluations. Green indicates agreement with the reference, red denotes deviations, orange marks missing ground-truth content, underlined text reflects template-style language, and blue highlights knowledge graph–grounded concepts.}
  \label{case2}
\end{figure}

To further illustrate the behavior of \texttt{PathMem}, we present a set of qualitative case studies as shown in \textcolor{red}{Figure~\ref{case12}, \textcolor{red}{Figure~\ref{KG-activationV2}, Figure~\ref{case1}, Figure~\ref{case3}}, Figure~\ref{case2}}.
These examples collectively provide an intuitive comparison across different methods and evaluation perspectives, highlighting variations in factual correctness, information completeness, and semantic alignment with the reference reports. Overall, the visualized results demonstrate that \texttt{PathMem} produces more accurate and comprehensive descriptions, while maintaining better consistency with domain knowledge and reducing template-like or spurious content compared to baseline approaches.

\FloatBarrier

\end{document}